\documentclass[11pt]{article}

\usepackage[final]{acl}

\usepackage{times}
\usepackage{latexsym}
\usepackage{color}

\usepackage[T1]{fontenc}

\usepackage[utf8]{inputenc}

\usepackage{microtype}

\usepackage{inconsolata}

\usepackage{graphicx}

\usepackage{multirow}
\usepackage{amssymb}  
\usepackage{amsmath}  
\usepackage{booktabs}
\usepackage{color}    
\usepackage{booktabs} 
\usepackage{pifont}

\usepackage{algorithm}
\usepackage{algpseudocode}
\definecolor{blue}{HTML}{4080E6}

\title{TimeSAF: Towards LLM-Guided Semantic Asynchronous Fusion for Time Series Forecasting}

\author{
	\textbf{Fan Zhang\textsuperscript{1}},
	\textbf{Shiming Fan\textsuperscript{1}},
	\textbf{Hua Wang\textsuperscript{2, \thanks{Corresponding author.}}},
	\\
	\\
	\textsuperscript{1}Shandong Technology and Business University,
	\textsuperscript{2}Ludong University
	\\
	\textbf{\tt \{zhangfan,2024410061\}@sdtbu.edu.cn,}
	\\
	\textbf{\tt hua.wang@ldu.edu.cn}
}

\begin{document}
\maketitle
\begin{abstract}
Despite the recent success of large language models (LLMs) in time-series forecasting, most existing methods still adopt a Deep Synchronous Fusion strategy, where dense interactions between textual and temporal features are enforced at every layer of the network. This design overlooks the inherent granularity mismatch between modalities and leads to what we term semantic perceptual dissonance: high-level abstract semantics provided by the LLM become inappropriately entangled with the low-level, fine-grained numerical dynamics of time series, making it difficult for semantic priors to effectively guide forecasting. To address this issue, we propose \textbf{TimeSAF}, a new framework based on hierarchical asynchronous fusion. Unlike synchronous approaches, TimeSAF explicitly decouples unimodal feature learning from cross-modal interaction. It introduces an independent cross-modal semantic fusion trunk, which uses learnable queries to aggregate global semantics from the temporal and prompt backbones in a bottom-up manner, and a stage-wise semantic refinement decoder that asynchronously injects these high-level signals back into the temporal backbone. This mechanism provides stable and efficient semantic guidance while avoiding interference with low-level temporal dynamics. Extensive experiments on standard long-term forecasting benchmarks show that TimeSAF significantly outperforms state-of-the-art baselines, and further exhibits strong generalization in both few-shot and zero-shot transfer settings.
\end{abstract}

\section{Introduction}

\begin{figure*}[h]   
	\begin{minipage}[b]{1\textwidth} 
		\includegraphics[width=\textwidth]{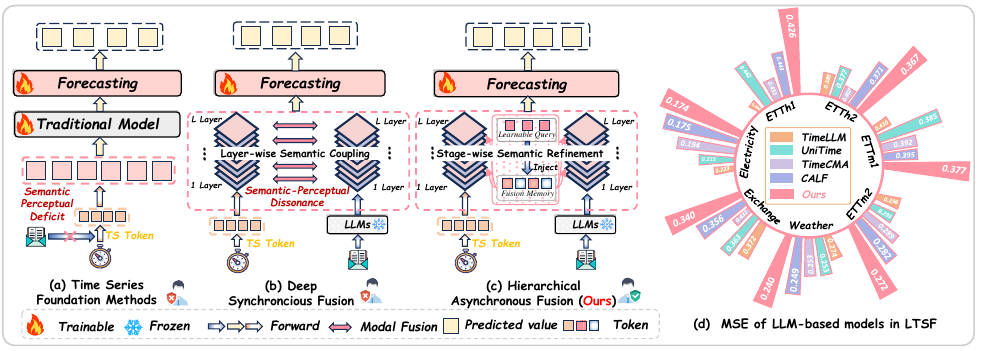}
		\centering
		
	\end{minipage}
	\caption{Comparison of strategy and performance between TimeSAF and other methods.}
	\label{fig:avgmse}
\end{figure*}

Long-term time series forecasting (LTSF) plays a crucial role in a wide range of real-world applications, including power load management \cite{fan2025cawformer,qiu2025dag}, traffic flow analysis \cite{2,shen2026mftformer,qiu2025DBLoss,wang2026idealtsf}, weather prediction \cite{liu2025rethinking}, and financial markets \cite{ariyo2014stock,zhang2026time}. Traditional time series analysis methods typically rely on statistical models or deep learning architectures to capture temporal dependencies from historical observations \cite{10,13,30,ma2025mofo,wang2026eeo}. However, as illustrated in Fig. \ref{fig:avgmse}(a), these models are often confined to the numerical modality and overlook the rich contextual information behind the series, such as metadata and event descriptions \cite{jin2023time,ge2025eventtsf}. This separation between numerical dynamics and semantic context leads to a \textbf{semantic perceptual deficit}, which limits the model’s ability to generalize across domains and makes it difficult to adapt to data-scarce scenarios \cite{zhao2025stem,ding2025dualsg} such as few-shot and zero-shot forecasting.

In recent years, large language models (LLMs) have been introduced into time-series forecasting to compensate for the lack of semantic priors, leveraging their strong reasoning ability and rich parametric knowledge. Existing LLM-based methods typically adopt the deep synchronous fusion strategy shown in Fig.~\ref{fig:avgmse}(b), i.e., layer-wise semantic coupling, where textual and temporal features are tightly aligned at every layer via dense cross-attention or feature concatenation \cite{wang2025freqllm,liu2024unitime}. However, this design ignores the \textbf{semantic perceptual dissonance} between discrete text and continuous time series: high-level abstract semantics are compressed into the same representational scale as low-level numerical fluctuations, leading to heavily entangled features that are hard to interpret or control. We term this effect semantic perceptual dissonance, where LLM priors cannot effectively guide temporal forecasting and may even cause negative transfer when fusion is performed at inappropriate depths.

To address these issues, we propose a hierarchical asynchronous fusion strategy, as illustrated in Fig. \ref{fig:avgmse}(c). Unlike designs that enforce synchronous interaction at every layer, the proposed scheme employs \textbf{stage-wise semantic refinement} to restrict cross-modal interaction to a few discrete stages: it first aggregates semantic representations from the time-series backbone and the prompt backbone in a bottom-up manner, and then injects these high-level semantics into deeper layers of the temporal backbone in a top-down manner. This asymmetric, cross-layer interaction prevents numerical modeling and semantic interaction from being entangled at all layers.

Building on this strategy, we develop TimeSAF, a multimodal time-series forecasting framework. Architecturally, TimeSAF constructs a compact semantic memory bank between temporal patterns and textual prompts via a fusion trunk parameterized by learnable queries, while embedding gated asynchronous refinement blocks into both unimodal backbones so that the temporal branch can selectively read from the fusion memory and update its representations. This design deliberately decouples feature extraction and multimodal fusion along the temporal depth, mitigating the interference caused by layer-wise synchronous fusion, while the fine-grained top-down refinement continuously aligns and injects task-relevant semantics from the fusion trunk into the numeric backbone. Extensive experiments show that TimeSAF achieves superior performance to existing methods across multiple LTSF benchmarks and multimodal settings. In summary, the contributions of this paper are as follows:
\begin{itemize}
	\item {\textbf{Fusion strategy.} We propose a hierarchical asynchronous fusion strategy that decouples unimodal encoding from cross-modal interaction, effectively alleviating the entanglement between numerical features and textual semantics.
	}
	\item {\textbf{Model architecture}. Building on this strategy, we introduce TimeSAF, which incorporates an independent cross-modal semantic fusion trunk and stage-wise semantic refinement decoder. The architecture first aggregates global semantics in a bottom-up manner and then injects the fused semantics back into the temporal backbone in a top-down fashion.}
	\item {\textbf{Empirical validation.} Extensive experiments on seven public benchmarks demonstrate that the proposed method consistently achieves state-of-the-art performance compared with both LLM-based and non-LLM baselines.}
\end{itemize}

\section{Related Work}\label{add:re}

With the rapid advances of deep learning  \cite{ENCODER,xiao2026points} in domains such as computer vision \cite{HABIT,INTENT}, video understanding \cite{HUD}, and multimodal representation learning \cite{xiaoreversible,ge2025t2s}, data-driven neural models have also become increasingly prevalent in time series forecasting.

Early time series forecasting predominantly relied on classical statistical models such as ARIMA, VAR, and STL with trend–seasonal decomposition \cite{siami2018comparison,schorfheide2005var,cleveland1990stl}. These approaches are robust in short-term and single-task settings, but are typically built on stationarity assumptions and are sensitive to high-dimensional nonlinear dependencies and complex noise. With the rise of deep learning, methods based on RNNs, CNNs, Transformers, and MLPs have become mainstream \cite{sherstinsky2020fundamentals,28,30}: RNN/LSTM/GRU model temporal dependencies via recurrent hidden states, convolutional architectures capture local temporal patterns and inter-variable relations, and Transformer-style models leverage global self-attention to better handle long-range dependencies \cite{ReTrack,OFFSET,REFINE,zhang2026decoding,fan2025fsmamba}. Building on this, PatchTST \cite{16} enhances long-sequence modeling through temporal patching and channel-independent design, while iTransformer \cite{15} and MoE/subspace-based methods \cite{qiu2025duet} mitigate multivariate heterogeneity and non-stationarity via channel reordering and pattern grouping. Despite this growing architectural diversity, these models still rely solely on historical numerical sequences to make deterministic predictions, which limits cross-domain generalization and leads to suboptimal performance in zero-shot and few-shot regimes.

More recently, large language models (LLMs) have been incorporated into time series forecasting \cite{gruver2023large,chang2025llm4ts}. Time-LLM \cite{jin2023time} adopts a time-series encoder with LLM interaction, jointly feeding encoded temporal features and textual prompts into the LLM so that it can assist trend understanding and pattern reasoning. GPT4TS \cite{zhou2023one} converts raw time series into token sequences via encoding, quantization, or descriptive prompts, and lets a pretrained LLM directly generate future trajectories. CALF \cite{liu2025calf} adopts a dual-stream architecture with dedicated loss functions to achieve deep cross-modal alignment; TimeCMA \cite{liu2025timecma} focuses on cross-channel alignment to alleviate channel entanglement; DualSG \cite{ding2025dualsg} employs a decoupled dual-stream design, where a numeric stream models fine-grained temporal dynamics and a semantic stream, driven by an LLM, performs trend-level semantic correction and channel-wise reasoning.

\textbf{Our Work.} Unlike prior LLM-based forecasters, TimeSAF first lets the temporal and semantic branches learn stable single-modal representations, and only then injects LLM-derived semantics back into the temporal backbone at a few staged levels, achieving a better balance between semantic guidance and robust numerical modeling.

\begin{figure*}[h]   
	\begin{minipage}[b]{1\textwidth} 
		\includegraphics[width=\textwidth]{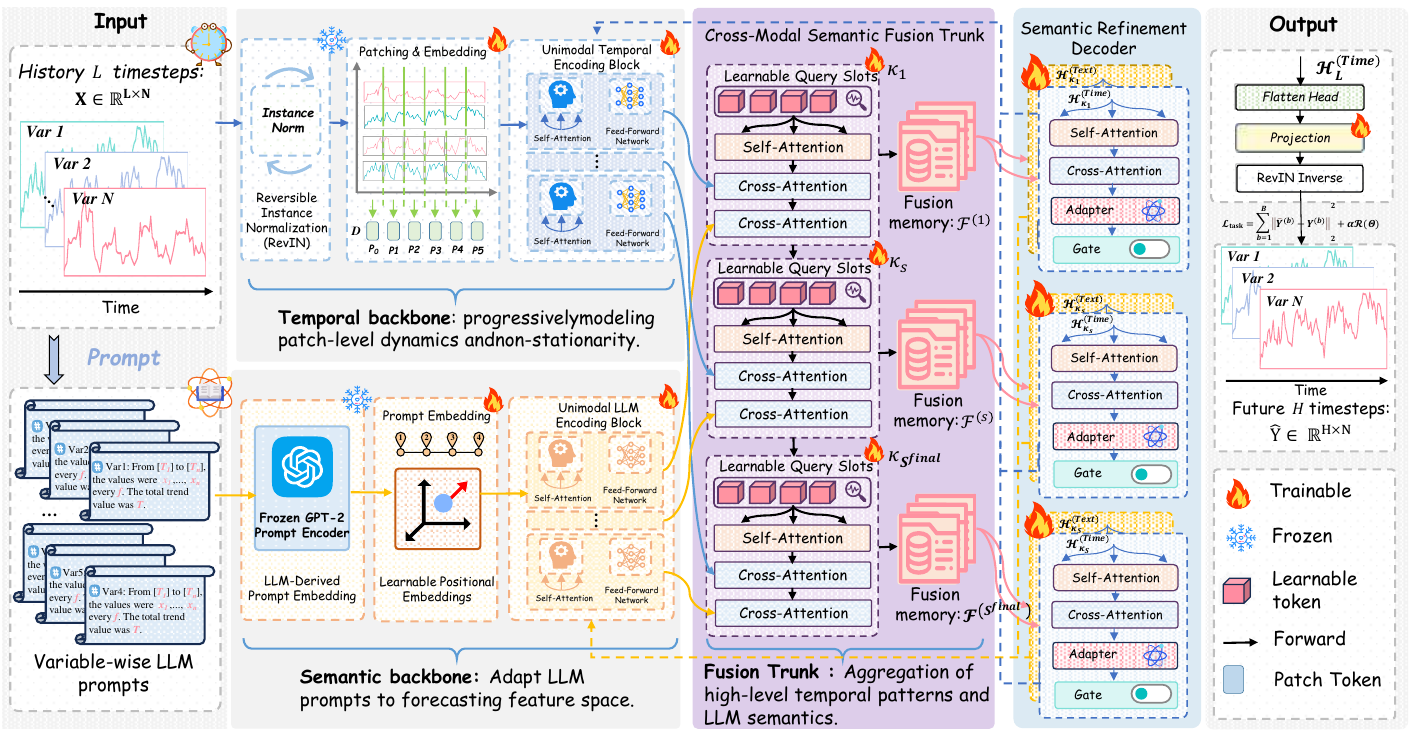}
		\centering
		
	\end{minipage}
	\caption{Overall architecture of TimeSAF.}
	\label{fig:model}
\end{figure*}

\section{Problem Formulation and Preliminary}
\subsection{Problem Formulation}
Given a multivariate time series \(\textbf{X} = \{ {x_1},...,{x_L}\}  \in {{\mathbb R}^{L \times N}}\) where \textit{L} denotes the length of the historical window and \textit{N} the number of variables, we further construct for each variable an offline LLM-derived prompt embedding \(\textbf{E} \in {{\mathbb R}^{{D_{{\rm{llm}}}} \times N}}\) where \({{D_{{\rm{llm}}}}}\) is the dimensionality of the textual feature space. Given an observation window \({{{\rm \textbf{X}}_{t - L + 1,t}}}\) and its associated prompts \(E\), the objective under a forecasting horizon of length \textit{H} is to learn a mapping:
\begin{equation}\label{eq:1}
	\begin{array}{l}
		{f_\theta }\left( {{{\rm \textbf{X}}_{t - L + 1,t}},\textbf{E}} \right) \mapsto {{\hat Y}_{t + 1:t + H}} \in {{\mathbb R}^{H \times N}}
	\end{array}
\end{equation}
\subsection{Time Series Encoding Branch}
In practical applications, time series often exhibit strong non-stationarity \cite{liu2024timebridge}. To alleviate this, we first apply reversible instance normalization (RevIN) \cite{kim2021reversible} to the input sequence. We then perform segmentation and embedding along the temporal dimension. Given a sequence of length $L$, with window length $P$ and stride $S$, the time axis is partitioned into 
$N_p = \left\lfloor \frac{L - P}{S} \right\rfloor + 1$ 
temporal patches, each containing $P$ consecutive time steps, forming the initial temporal patch tokens
$\mathbf{P}^{Time}_{t-L+1,t} \in \mathbb{R}^{N_p \times P}$:
\begin{equation}\label{eq:2}
	\mathbf{P}^{Time}_{t-L+1,t} = \text{Patching}(\mathbf{X}_{t-L+1,t}).
\end{equation}
A projection layer $g(\cdot)$ is then applied along the temporal dimension to map each patch to a $D$-dimensional representation. With an added learnable positional embedding $e^{pos}$, we obtain the final temporal tokens, which are used as input to the subsequent encoder:
\begin{equation}\label{eq:3}
	\mathbf{X}^{time}_{i,t-L+1,t} = g(\mathbf{P}_{i,t-L+1,t}) + e^{pos}.
\end{equation}

\subsection{LLM-based Prompt Encoding Branch}
To inject external priors and semantic structure into the model, we introduce an LLM-based prompt encoding branch. This branch leverages a pretrained and frozen GPT-2 backbone to process natural language descriptions associated with segments of the input sequence. Following \cite{liu2025timecma}, we automatically generate, for each variable in $\mathbf{X}$, a prompt describing its statistics over the observation window (see Appendix~\ref{app:llm} for details). Each prompt is tokenized by the GPT-2 tokenizer and fed into the frozen GPT-2 encoder, yielding prompt representations $\mathbf{E} \in \mathbb{R}^{D_{\text{llm}} \times N}$, where $N$ is the number of variables and $D_{\text{llm}}$ is the LLM embedding dimension (768 for GPT-2). To ensure consistency within a mini-batch, all prompt sequences are padded to a unified length. We then introduce a learnable semantic adaptation module $l(\cdot)$ that maps $\mathbf{E}$ from the original LLM embedding space to the model semantic space $\mathbb{R}^D$, producing node-wise semantic features. A learnable positional embedding $e$ is further added to each variable, resulting in the textual modality embedding $\mathbf{X}^{Text} \in \mathbb{R}^{D \times N}$, which serves as the input to the subsequent semantic encoding branch.

\section{Overall Architecture of TimeSAF}
In this section, we outline the overall architecture of \textbf{TimeSAF}. As illustrated in Fig.~\ref{fig:model}, given a multivariate historical time series and its associated LLM-based prompts, TimeSAF first encodes the numeric input with a patching-based temporal encoder, while a frozen GPT-2 backbone produces prompt-derived semantic representations. A semantic fusion trunk is inserted at several predefined layers, where a set of learnable queries aggregates information from both the temporal and semantic branches. Subsequently, asynchronous refinement modules inject the fused semantics back into the temporal backbone, and a lightweight prediction head maps the refined temporal tokens to future forecasts.

\subsection{Unimodal Encoding Backbones}
After obtaining the encoded time-series tokens and LLM-based prompt embeddings, we construct two structurally symmetric unimodal backbones for the numerical and semantic modalities, respectively. Both backbones are composed of several stacked Unimodal Encoding blocks, each consisting of a self-attention layer and a feed-forward network, which progressively extract higher-level representations within each modality without introducing any cross-modal interaction. Let \({{\cal H}_l}\) denote the input to the \textit{l}-th layer of a unimodal backbone:
\begin{equation}\label{eq:4}
	\begin{array}{l}
		{{\cal {U}}_l} = {{\cal {H}}_l} + SelfAttn({{\cal {H}}_l})
	\end{array}
\end{equation}
Then, a position-wise feed-forward network further transforms the intermediate representation to produce the output of the (\textit{l}+1)-th layer.
\begin{equation}\label{eq:5}
	\begin{array}{l}
		{{\cal H}_{l + 1}} = {{\cal U}_l} + FFN({{\cal U}_l})
	\end{array}
\end{equation}
Here, \(SelfAttn(\cdot)\) denotes a multi-head self-attention module with an internal layer normalization, and \(FFN(\cdot)\)  denotes a position-wise feed-forward network. Stacking multiple Unimodal Encoding blocks on the temporal modality allows the model to progressively enrich the latent representation of the series at the patch level, while on the textual modality, the LLM-derived semantic vectors are gradually adapted to the feature space of the backbone network. It is worth noting that, at this stage, the two backbones evolve strictly within their own modalities, providing stable and semantically well-formed unimodal representations for subsequent cross-modal fusion.
\subsection{Cross-Modal Semantic Fusion Trunk}

After the layer-wise encoding of the Unimodal Encoding Backbones, we obtain high-level representations for both the temporal and textual modalities. We then introduce an independent \textbf{Cross-Modal Semantic Fusion Trunk}, which performs explicit cross-modal aggregation at a set of designated fusion layers, and compresses information from both branches into a fixed-length fusion memory.

Formally, each unimodal backbone contains $dp$ blocks, and we predefine $S$ fusion stages, yielding $L_S = dp / S$ depth intervals. Let $\kappa_s$ be the layer index where the $s$-th fusion stage is triggered. At the beginning of this stage, we instantiate a set of learnable fusion queries
${\cal Q}_s^F \in \mathbb{R}^{P_f \times D_f}$, which are broadcast across samples and variables during the forward pass to form the initial fusion representation
${\cal H}_{s,0}^F \in \mathbb{R}^{(BN) \times P_f \times D_f}$. When the temporal and textual backbones reach layer $\kappa_s$, their hidden states
${\cal H}_{\kappa_s}^{Time}$ and ${\cal H}_{\kappa_s}^{Text}$ are used as key–value inputs, and a bottom–up semantic aggregation is performed. We first apply self-attention over the fusion queries:
\begin{equation}\label{eq:6}
	\tilde{\cal H}_s^F = {\cal H}_{s,0}^F + SelfAttn({\cal H}_{s,0}^F).
\end{equation}
Then, taking $\tilde{\cal H}_s^F$ as queries, we retrieve information from the temporal and textual branches via two cross-attention operations:
\begin{equation}\label{eq:7}
	\begin{aligned}
		\tilde{\cal H}_s^F &\leftarrow \tilde{\cal H}_s^F + CrossAttn_{Time}(\tilde{\cal H}_s^F, {\cal H}_{\kappa_s}^{Time}),\\
		\tilde{\cal H}_s^F &\leftarrow \tilde{\cal H}_s^F + CrossAttn_{Text}(\tilde{\cal H}_s^F, {\cal H}_{\kappa_s}^{Text}).
	\end{aligned}
\end{equation}
Finally, a position-wise feed-forward network integrates these signals and yields the fusion memory for the $s$-th stage:
\begin{equation}\label{eq:8}
	{\cal F}^{(s)} = \tilde{\cal H}_s^F + FFN(\tilde{\cal H}_s^F).
\end{equation}

By repeating this procedure over all $S$ stages, the model obtains a set of fusion memories
$\{ {\cal F}^{(1)}, \dots, {\cal F}^{(S)} \}$ that compactly encode high-level semantic information from both backbones. Cross-modal interactions occur only at the designated fusion layers, while the temporal and semantic branches evolve independently in the remaining layers. This stage-wise design allows sufficient multi-modal integration, while avoiding the heavy coupling and optimization instability that arise when cross-attention is inserted at every layer, and provides a clean contextual interface for the subsequent asynchronous refinement modules. A simplified theoretical analysis under linearized assumptions is provided in Appendix~\ref{app:theory} to further support the intuition behind this design.

\begin{table*}[!ht]   
	\caption{Average MSE and MAE over four prediction lengths. All experiments fix the lookback length \textit{T} = 96. The prediction length set is \textit{H} $ \in $ \{96, 192, 336, 720\}. The best result is \textcolor{red}{\textbf{red}}, the second best result is \textcolor{blue}{\underline{underlined}}. Our full results are in Appendix \ref{sec:full}.}
	\label{tab:allmse}
	\centering
	\renewcommand{\arraystretch}{1.5}
	\tiny
	\setlength{\tabcolsep}{2.5pt}
	\begin{tabular}{c|cc|cc|cc|cc|cc|cc|cc|cc|cc|cc|cc}
		\toprule[1.2pt]
		
		\multirow{1}{*}{\textbf{Models}} 
		& \multicolumn{2}{c|}{\textbf{TimeSAF}} 
		& \multicolumn{2}{c|}{\textbf{CALF}} 
		& \multicolumn{2}{c|}{\textbf{TimeCMA}} 
		& \multicolumn{2}{c|}{\textbf{Time-FFM}}
		& \multicolumn{2}{c|}{\textbf{UniTime}} 
		& \multicolumn{2}{c}{\textbf{Time-LLM}}  
		& \multicolumn{2}{c|}{\textbf{GPT4TS}} 
		& \multicolumn{2}{c|}{\textbf{iTransformer}} 
		& \multicolumn{2}{c|}{\textbf{PatchTST}} 
		& \multicolumn{2}{c|}{\textbf{Crossformer}} 
		& \multicolumn{2}{c}{\textbf{FEDformer}} \\
		
		\cmidrule(r){1-23}
		\multirow{1}{*}{\textbf{Metrics}} 
		& \textbf{MSE} & \multicolumn{1}{c|}{\textbf{MAE}} 
		& \textbf{MSE} & \multicolumn{1}{c|}{\textbf{MAE}}
		& \textbf{MSE} & \multicolumn{1}{c|}{\textbf{MAE}}
		& \textbf{MSE} & \multicolumn{1}{c|}{\textbf{MAE}}
		& \textbf{MSE} & \multicolumn{1}{c|}{\textbf{MAE}} 
		& \textbf{MSE} & \multicolumn{1}{c|}{\textbf{MAE}} 
		& \textbf{MSE} & \multicolumn{1}{c|}{\textbf{MAE}} 
		& \textbf{MSE} & \multicolumn{1}{c|}{\textbf{MAE}} 
		& \textbf{MSE} & \multicolumn{1}{c|}{\textbf{MAE}}
		& \textbf{MSE} & \multicolumn{1}{c|}{\textbf{MAE}} 
		& \textbf{MSE} & \multicolumn{1}{c}{\textbf{MAE}}\\
		\midrule
		
		\textbf{ETTm1} 
		& \textbf{\textcolor{red}{0.377}} & \textbf{\textcolor{red}{0.390}} 
		& 0.395 & \textcolor{blue}{\underline{0.390}} 
		& 0.392 & 0.405 
		& 0.399 & 0.402
		& \textcolor{blue}{\underline{0.385}} & 0.399 
		& 0.410 & 0.409 
		& 0.396 & 0.400 
		& 0.407 & 0.410 
		& 0.387 & 0.400 
		& 0.502 & 0.502 
		& 0.448 & 0.452 \\
		\midrule
		
		\textbf{ETTm2} 
		& \textbf{\textcolor{red}{0.272}} & \textbf{\textcolor{red}{0.320}} 
		& 0.282 & \textcolor{blue}{\underline{0.321}} 
		& 0.289 & 0.332 
		& 0.286 & 0.332
		& 0.293 & 0.334 
		& 0.296 & 0.340 
		& 0.293 & 0.338 
		& 0.288 & 0.332 
		& \textcolor{blue}{\underline{0.281}} & 0.326 
		& 1.216 & 0.707 
		& 0.304 & 0.349 \\
		\midrule
		
		\textbf{ETTh1} 
		& \textbf{\textcolor{red}{0.426}} & \textbf{\textcolor{red}{0.428}} 
		& 0.443 & 0.435 
		& 0.451 & 0.449 
		& 0.442 & \textcolor{blue}{\underline{0.434}}
		& 0.442 & 0.447 
		& 0.460 & 0.449 
		& 0.456 & 0.450 
		& 0.454 & 0.447 
		& 0.469 & 0.454 
		& 0.620 & 0.572 
		& \textcolor{blue}{\underline{0.440}} & 0.459 \\
		\midrule
		
		\textbf{ETTh2} 
		& \textbf{\textcolor{red}{0.367}} & \textbf{\textcolor{red}{0.394}} 
		& \textcolor{blue}{\underline{0.371}} & \textcolor{blue}{\underline{0.394}} 
		& 0.407 & 0.419 
		& 0.382 & 0.406
		& 0.377 & 0.402 
		& 0.389 & 0.408 
		& 0.456 & 0.413 
		& 0.383 & 0.407 
		& 0.387 & 0.407 
		& 0.942 & 0.683 
		& 0.436 & 0.449 \\
		\midrule
		
		\textbf{Weather} 
		& \textbf{\textcolor{red}{0.240}} & \textbf{\textcolor{red}{0.268}} 
		& \textcolor{blue}{\underline{0.249}} & \textcolor{blue}{\underline{0.273}} 
		& 0.253 & 0.283 
		& 0.270 & 0.288
		& 0.253 & 0.276 
		& 0.274 & 0.290 
		& 0.278 & 0.297 
		& 0.257 & 0.278 
		& 0.259 & 0.281 
		& 0.258 & 0.315 
		& 0.308 & 0.360 \\
		\midrule
		
		\textbf{Exchange} 
		& \textcolor{blue}{\underline{0.340}} & \textbf{\textcolor{red}{0.385}} 
		& 0.356 & 0.402
		& 0.432 & 0.446 
		& \textbf{\textcolor{red}{0.338}} & \textcolor{blue}{\underline{0.391}}
		& 0.363 & 0.404 
		& 0.372 & 0.416 
		& 0.370 & 0.409 
		& 0.360 & 0.403 
		& 0.367 & 0.404 
		& 0.470 & 0.477 
		& 0.518 & 0.428 \\
		\midrule
		
		\textbf{Electricity} 
		& \textbf{\textcolor{red}{0.174}} & \textbf{\textcolor{red}{0.264}} 
		& \textcolor{blue}{\underline{0.175}} & \textcolor{blue}{\underline{0.265}} 
		& 0.194 & 0.287 
		& 0.216 & 0.299
		
		& 0.215 & 0.304 
		& 0.223 & 0.309 
		& 0.217 & 0.307 
		& 0.178 & 0.270 
		& 0.205 & 0.290 
		& 0.244 & 0.334 
		& 0.213 & 0.326 \\	
		
		\bottomrule[1.2pt]
	\end{tabular}
	\vspace{0.5em}
\end{table*}
\subsection{Semantic Refinement Decoder}
After obtaining the stage-wise fusion memory \({\cal F}^{(s)}\) from the semantic fusion trunk, the Semantic Refinement Decoder feeds this high-level semantic context back into both unimodal backbones. Let \(m \in \{Time, Text\}\) index the temporal and textual modalities, and denote by \({\cal H}_l^{(m)}\) the input to the \(l\)-th layer of modality \(m\). Before cross-modal refinement, we first perform an intra-modal self-attention update:
\begin{equation}\label{eq:9}
	{\cal U}_l^{(m)} = {\cal H}_l^{(m)} + \mathrm{SelfAttn}^{(m)}\!\big({\cal H}_l^{(m)}\big),
\end{equation}
where \(\mathrm{SelfAttn}^{(m)}\) is a multi-head self-attention block with layer normalization.

The intermediate representation \({\cal U}_l^{(m)}\) is then refined using the fusion memory \({\cal F}^{(s)}\) via cross-attention:
\begin{equation}\label{eq:10}
	{\cal Z}_l^{(m)} = \mathrm{CrossAttn}^{(m)}\!\big({\cal U}_l^{(m)}, {\cal F}^{(s)}\big),
\end{equation}
where \(\mathrm{CrossAttn}^{(m)}(\cdot,\cdot)\) shares the same form as the cross-attention used in the Fusion Block and treats \({\cal F}^{(s)}\) as shared key–value context.

To obtain modality-specific refinement while keeping the injection strength controllable, each modality is equipped with an independent linear adapter \({\cal W}_{ad}^{(m)}\), and a scalar gate \(g \in \mathbb{R}\) shared across modalities. The gated refinement residual is
\begin{equation}\label{eq:11}
	{\cal R}_l^{(m)} = \sigma(g)\,{\cal W}_{ad}^{(m)}\!\big({\cal Z}_l^{(m)}\big),
\end{equation}
where \(\sigma(\cdot)\) is the sigmoid function, and \({\cal R}_l^{(m)}\) represents the refined signal contributed by the fusion memory. Finally, we add this residual back to the intra-modal representation and apply a position-wise feed-forward network to obtain the layer output:
\begin{equation}\label{eq:12}
	\begin{array}{l}
			\hat {\cal H}_l^{(m)} = {{\cal U}_l}^{(m)} + {\cal R}_l^{(m)}\\
			{\cal H}_{l + 1}^{(m)} = \hat {\cal H}_l^{(m)} + FF{N^{(m)}}(\hat {\cal H}_l^{(m)})
		\end{array}
\end{equation}

Architecturally, the temporal and textual branches share the same fusion memory \({\cal F}^{(s)}\) inside the Semantic Refinement Decoder, but project it back to their own representation spaces through separate adapters \({\cal W}_{ad}^{(Time)}\) and \({\cal W}_{ad}^{(Text)}\). This shared but modality-specific refinement drives the two branches toward a compatible latent subspace. At the beginning layer \(\kappa_s\) of each fusion stage, a new fusion memory \({\cal F}^{(s)}\) is computed by the Cross-Modal Semantic Fusion Trunk and then reused within that stage to update \({\cal H}_l^{Time}\) and \({\cal H}_l^{Text}\) via \({\cal R}_l^{Time}\) and \({\cal R}_l^{Text}\), respectively, until the next stage produces a new fusion memory that replaces it as the contextual signal.

\subsection{Output Projection and Optimization Objective}

Finally, the last-layer temporal representation ${\cal H}_dp^{Time}$ is fed into a linear output head, which first flattens the patch-wise features and then maps them to the prediction horizon:
\begin{equation}\label{eq:13}
	\mathbf{Y} = Flatten({\cal H}_dp^{Time}) W_{\text{out}} + \mathbf{b}_{\text{out}}.
\end{equation}
The resulting forecast $\mathbf{Y}$ is then de-normalized via the inverse RevIN operation to obtain $\hat{\mathbf{Y}} \in \mathbb{R}^{H \times N}$ on the original value scale. TimeSAF is trained with a mean squared error loss plus standard $\ell_2$ weight decay. Let $\Theta$ denote all trainable parameters; the overall objective is
\begin{equation}\label{eq:14}
	{\cal L} = {\cal L}_{\text{pred}} + \alpha {\cal R}(\Theta),
\end{equation}
where $\alpha \ge 0$ is a balancing coefficient, and
\begin{equation}\label{eq:15}
	{\cal L}_{\text{pred}} = \frac{1}{B} \sum_{b=1}^B \left\| \hat{\mathbf{Y}}^{(b)} - \mathbf{Y}^{(b)} \right\|_2^2,
	\quad
	{\cal R}(\Theta) = \sum_{\theta \in \Theta} \theta^2
\end{equation}
The objective ${\cal L}$ is minimized by back-propagation.

\begin{table*}[!ht]  
	\caption{Few-shot forecasting performance on ETT datasets using only 10\% of the training data. All experiments fix the lookback length \textit{T} = 96. The prediction horizon is set to \textit{H} \( \in \) \{96, 192, 336, 720\}. Our full results are in Appendix \ref{sec:full}.}
	\label{tab:10}
	\centering
	\renewcommand{\arraystretch}{1.5}
	\tiny 
	\setlength{\tabcolsep}{2.7pt} 
	\begin{tabular}{c|cc|cc|cc|cc|cc|cc|cc|cc|cc|cc|cc}
		\toprule[1.2pt]
		
		\multirow{1}{*}{\textbf{Models}}  & \multicolumn{2}{c|}{\textbf{TimeSAF}} & \multicolumn{2}{c|}{\textbf{CALF}} & \multicolumn{2}{c|}{\textbf{TimeCMA}} & \multicolumn{2}{c|}{\textbf{Time-LLM }} & \multicolumn{2}{c|}{\textbf{GPT4TS}}  & \multicolumn{2}{c|}{\textbf{PatchTST}} & \multicolumn{2}{c|}{\textbf{Crossformer}} & \multicolumn{2}{c|}{\textbf{FEDformer}} & \multicolumn{2}{c|}{\textbf{TimesNet}}  & \multicolumn{2}{c|}{\textbf{DLinear}} & \multicolumn{2}{c}{\textbf{MICN}} \\
		
		\cmidrule(r){1-23}
		\multirow{1}{*}{\textbf{Metrics}} 
		& \textbf{MSE} & \multicolumn{1}{c|}{\textbf{MAE}} & \textbf{MSE}  & \multicolumn{1}{c|}{\textbf{MAE}}& \textbf{MSE} & \multicolumn{1}{c|}{\textbf{MAE}}& \textbf{MSE}  & \multicolumn{1}{c|}{\textbf{MAE}} & \textbf{MSE}  & \multicolumn{1}{c|}{\textbf{MAE}} & \textbf{MSE} & \multicolumn{1}{c|}{\textbf{MAE}} & \textbf{MSE}  & \multicolumn{1}{c|}{\textbf{MAE}} & \textbf{MSE} & \multicolumn{1}{c|}{\textbf{MAE}}& \textbf{MSE}  & \multicolumn{1}{c|}{\textbf{MAE}} & \textbf{MSE}  & \multicolumn{1}{c|}{\textbf{MAE}}&\textbf{MSE} & \multicolumn{1}{c}{\textbf{MAE}}\\
		\midrule
		
		\textbf{ETTm1} 
		& \textbf{\textcolor{red}{0.483}} & \textbf{\textcolor{red}{0.449}} &  \textcolor{blue}{\underline{0.504}} &  \textcolor{blue}{\underline{0.462}} & 0.594 & 0.504 & 0.636 & 0.512 & 0.608 & 0.500 & 0.557 & 0.483 & 1.340 & 0.848 & 0.696 & 0.572 & 0.673 & 0.534 & 0.567 & 0.499 & 0.970 & 0.674 \\
		\midrule
		\textbf{ETTm2} 
		& \textbf{\textcolor{red}{0.294}} & \textbf{\textcolor{red}{0.330}} & 0.302 &  \textcolor{blue}{\underline{0.330}} & 0.309 & 0.344 & 0.308 & 0.343 & 0.303 & 0.336 &  \textcolor{blue}{\underline{0.295}} & 0.334 & 1.985 & 1.048 & 0.356 & 0.392 & 0.321 & 0.354 & 0.329 & 0.382 & 1.073 & 0.716 \\
		\midrule
		\textbf{ETTh1} 
		& \textbf{\textcolor{red}{0.622}} & \textbf{\textcolor{red}{0.527}} &  \textcolor{blue}{\underline{0.644}} &  \textcolor{blue}{\underline{0.541}} & 0.721 & 0.575 & 0.765 & 0.584 & 0.689 & 0.555 & 0.683 & 0.645 & 1.744 & 0.914 & 0.750 & 0.607 & 0.865 & 0.625 & 0.647 & 0.552 & 1.405 & 0.814 \\
		\midrule
		\textbf{ETTh2} 
		&  \textcolor{blue}{\underline{0.422}} & \textbf{\textcolor{red}{0.423}} & \textbf{\textcolor{red}{0.419}} &  \textcolor{blue}{\underline{0.427}} & 0.451 & 0.448 & 0.589 & 0.498 & 0.579 & 0.497 & 0.550 & 0.487 & 3.139 & 1.378 & 0.553 & 0.525 & 0.476 & 0.463 & 0.441 & 0.458 & 2.533 & 1.158 \\
		
		\bottomrule[1.2pt]
	\end{tabular}
	\vspace{0.5em}
\end{table*}

\begin{table*}[!ht]   
	\caption{Zero-shot forecasting performance on the ETT datasets, where prediction lengths $H \in \{96,192,336,720\}$. “h1→m1” indicates that models trained on ETTh1 are evaluated on ETTm1, and similarly for the other transfer
		settings. Our full results are in Appendix \ref{sec:full}.}
	\label{tab:0}
	\centering
	\renewcommand{\arraystretch}{1.5}
	\tiny 
	\setlength{\tabcolsep}{2.5pt} 
	\begin{tabular}{c|cc|cc|cc|cc|cc|cc|cc|cc|cc|cc|cc}
		\toprule[1.2pt]
		
		\multirow{1}{*}{\textbf{Models}}  & \multicolumn{2}{c|}{\textbf{TimeSAF}} & \multicolumn{2}{c|}{\textbf{CALF}} & \multicolumn{2}{c|}{\textbf{TimeCMA}} & \multicolumn{2}{c|}{\textbf{Time-LLM }} & \multicolumn{2}{c|}{\textbf{GPT4TS}}  & \multicolumn{2}{c|}{\textbf{PatchTST}} & \multicolumn{2}{c|}{\textbf{Crossformer}} & \multicolumn{2}{c|}{\textbf{FEDformer}} & \multicolumn{2}{c|}{\textbf{TimesNet}}  & \multicolumn{2}{c|}{\textbf{DLinear}} & \multicolumn{2}{c}{\textbf{MICN}} \\
		
		\cmidrule(r){1-23}
		\multirow{1}{*}{\textbf{Metrics}} 
		& \textbf{MSE} & \multicolumn{1}{c|}{\textbf{MAE}} & \textbf{MSE}  & \multicolumn{1}{c|}{\textbf{MAE}}& \textbf{MSE} & \multicolumn{1}{c|}{\textbf{MAE}}& \textbf{MSE}  & \multicolumn{1}{c|}{\textbf{MAE}} & \textbf{MSE}  & \multicolumn{1}{c|}{\textbf{MAE}} & \textbf{MSE} & \multicolumn{1}{c|}{\textbf{MAE}} & \textbf{MSE}  & \multicolumn{1}{c|}{\textbf{MAE}} & \textbf{MSE} & \multicolumn{1}{c|}{\textbf{MAE}}& \textbf{MSE}  & \multicolumn{1}{c|}{\textbf{MAE}} & \textbf{MSE}  & \multicolumn{1}{c|}{\textbf{MAE}}&\textbf{MSE} & \multicolumn{1}{c}{\textbf{MAE}}\\
		\midrule

		\textbf{h1 → m1} 
		& \textbf{\textcolor{red}{0.749}} & \textbf{\textcolor{red}{0.563}} &  \textcolor{blue}{\underline{0.755}} &  \textcolor{blue}{\underline{0.574}} & 0.820 & 0.590 & 0.847 & 0.600 & 0.798 & 0.574 & 0.894 & 0.610 & 0.999 & 0.736 & 0.765 & 0.588 & 0.794 & 0.575 & 1.439 & 0.870 & 0.760 & 0.577 \\	
		
		\midrule
		
		\textbf{h1 → m2} 
		& \textbf{\textcolor{red}{0.313}} & \textbf{\textcolor{red}{0.355}} & 0.316 &  \textcolor{blue}{\underline{0.355}} & 0.329 & 0.365 &  \textcolor{blue}{\underline{0.315}} & 0.357 & 0.317 & 0.359 & 0.318 & 0.362 & 1.120 & 0.789 & 0.357 & 0.403 & 0.339 & 0.370 & 2.428 & 1.236 & 0.399 & 0.439 \\
		
		\midrule
		
		\textbf{h2 → m1}
		& 0.873 & 0.604 & 0.836 & \textbf{\textcolor{red}{0.586}} & 1.087 & 0.673 & 0.868 & 0.595 & 0.920 & 0.610 & 0.871 & 0.596 & 1.195 & 0.711 & \textbf{\textcolor{red}{0.741}} &  \textcolor{blue}{\underline{0.588}} & 1.286 & 0.705 &  \textcolor{blue}{\underline{0.764}} & 0.601 & 0.778 & 0.594 \\
		
		\midrule
		
		\textbf{h2 → m2}
		& \textbf{\textcolor{red}{0.317}} & \textbf{\textcolor{red}{0.360}} &  \textcolor{blue}{\underline{0.319}} &  \textcolor{blue}{\underline{0.360}} & 0.349 & 0.386 & 0.322 & 0.363 & 0.331 & 0.371 & 0.420 & 0.433 & 2.043 & 1.124 & 0.365 & 0.405 & 0.361 & 0.390 & 0.527 & 0.519 & 0.496 & 0.496 \\
		
		\bottomrule[1.2pt]
	\end{tabular}
	\vspace{0.5em}
\end{table*}
\section{Experiments}
\textbf{Datasets and Metrics.} We evaluate TimeSAF on seven widely used multivariate time series benchmarks: the four subsets of the Electricity Transformer Temperature (ETT) dataset (ETTh1, ETTh2, ETTm1, and ETTm2), together with Electricity, Weather, and Exchange. In line with standard practice in forecasting studies, we adopt Mean Absolute Error (MAE) and Mean Squared Error (MSE) as our primary evaluation metrics. The detailed statistics of these datasets are summarized in Appendix \ref{app:datasets}.

\textbf{Baselines.} We compare TimeSAF against a diverse set of recent and representative forecasting models.
(1) LLM-based models: CALF \cite{liu2025calf}, TimeCMA \cite{liu2025timecma},Time-FFM \cite{liu2024time} , UniTime \cite{liu2024unitime}, Time-LLM \cite{jin2023time}, and GPT4TS \cite{zhou2023one}.
(2) Transformer-based models: iTransformer\cite{15}, PatchTST \cite{16}, Crossformer \cite{25}, and FEDformer \cite{9}.
(3) CNN-based models: TimesNet \cite{28} and MICN \cite{wang2023micn}.
(4) MLP-based models: DLinear \cite{10}.

\textbf{Implementation Details.} We conduct all experiments under a unified evaluation pipeline and adopt the same configuration as \cite{28} to ensure a fair comparison with strong baselines.  We use a pretrained GPT-2 model (the first six layers) \cite{28} as the default LLM backbone. TimeSAF is optimized using the Adam optimizer, trained for up to 50 epochs with early stopping. All experiments are run on 8 NVIDIA GeForce RTX 3090 GPUs (24 GB each). See Appendix \ref{app:details} for more details.

\subsection{Long-term Forecasting}
\textbf{Setups.} For a fair comparison, we fix the input sequence length to \textit{L} = 96 and consider four forecasting horizons \(H \in \{ 96,192,336,720\} \). Consistent with the TFB-based setup \cite{xu2023fits}, we maintain the same configuration but do not use the “Drop-Last” trick during training to ensure fair comparison.

\textbf{Results.} The overall results are summarized in Table \ref{tab:allmse}. On the aggregated comparison across all datasets and forecasting horizons, TimeSAF consistently outperforms all baselines and achieves the best performance on all 14 aggregated metrics. Compared with state-of-the-art LLM-based methods (CALF, TimeCMA, Time-FFM, UniTime, and Time-LLM), TimeSAF reduces MSE by 3.10\%, 8.83\%, 6.58\%, and 10.31\%, respectively. It also clearly surpasses Transformer-, CNN-, and MLP-based models, with improvements over these baselines typically exceeding 5\%. These experimental results demonstrate that our TimeSAF can fully utilize the temporal patterns and semantic information in a limited input sequence, thus enabling accurate predictions.

\subsection{Few/zero-shot Learning}
\textbf{Setups.} Given that large language models (LLMs) have demonstrated strong generalization in few-shot and zero-shot learning settings \cite{zhou2023one,jin2023time}, this property is particularly relevant for real-world time series forecasting under data-scarce scenarios. Therefore, we also evaluate the performance of TimeSAF in few-shot and zero-shot regimes. In the few-shot setting, we use the ETT datasets and restrict the training data to only 10\% of the original training split. In the zero-shot setting, the model trained on one dataset is directly deployed to a new, unseen dataset without any additional training, while keeping the training hyperparameters identical to those used in the long-term forecasting experiments.

\textbf{Few-shot Learning.}
Table \ref{tab:10} reports the results on the challenging few-shot forecasting task under different prediction horizons. With only a small fraction of training samples, TimeSAF achieves the best performance on 7 out of 8 overall metrics, indicating that the model can effectively extract useful patterns from limited historical data. Compared with LLM-based baselines (CALF, TimeCMA, Time-LLM, and GPT4TS), TimeSAF achieves relative MSE improvements of 2.38\%, 10.93\%, 18.91\%, and 15.09\%, respectively. In addition, TimeSAF outperforms the strong Transformer-based baseline PatchTST by 11.46\%, further confirming its advantage in the few-shot setting.

\textbf{Zero-shot Learning.} To further assess the cross-dataset generalization ability of TimeSAF, we conduct zero-shot transfer experiments, where the model is trained on one dataset and then directly evaluated on a different dataset without any additional fine-tuning. As reported in Table \ref{tab:0}, TimeSAF achieves the lowest average MSE on three out of four transfer directions. Overall, its zero-shot performance is comparable to the strongest LLM-based baseline, CALF, while consistently outperforming TimeCMA, Time-LLM, and GPT4TS, with average MSE reductions of 10.60\%, 3.29\%, and 4.19\%, respectively. These results indicate that the proposed asynchronous fusion framework offers highly competitive zero-shot transfer capability compared with existing LLM-enhanced forecasting models.
\begin{figure}[h]   
	\begin{minipage}[b]{0.48\textwidth} 
		\includegraphics[width=\textwidth]{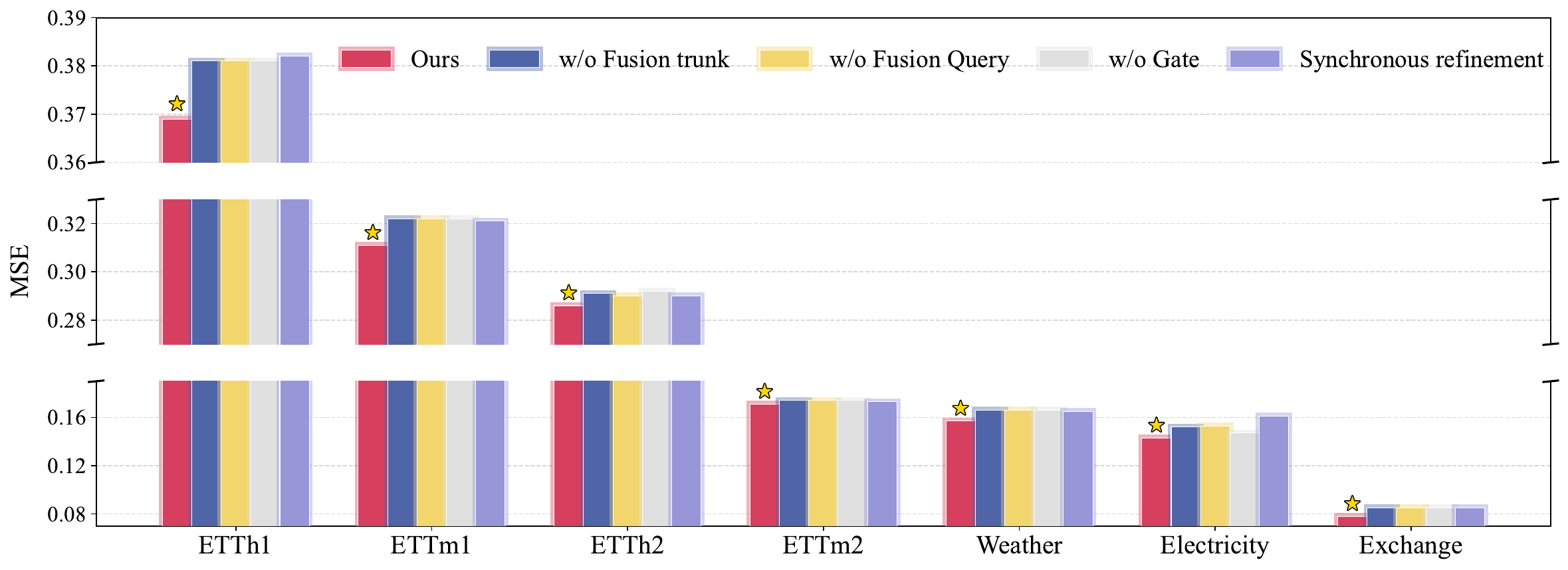}
		\centering
		
	\end{minipage}
	\caption{Ablation studies of different variants of TimeSAF on multiple datasets.}
	\label{fig:ab}
\end{figure}

\begin{figure*}[!ht]  
	\centering
	\begin{minipage}[b]{0.24\textwidth}
		\includegraphics[width=\textwidth]{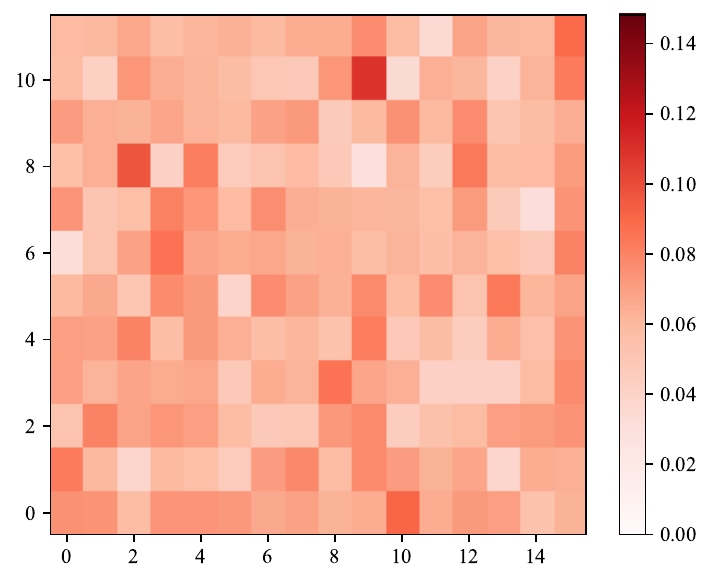}
		\centering
		\text{( \textit{a} )}
	\end{minipage}
	\hfill
	\begin{minipage}[b]{0.24\textwidth}
		\includegraphics[width=\textwidth]{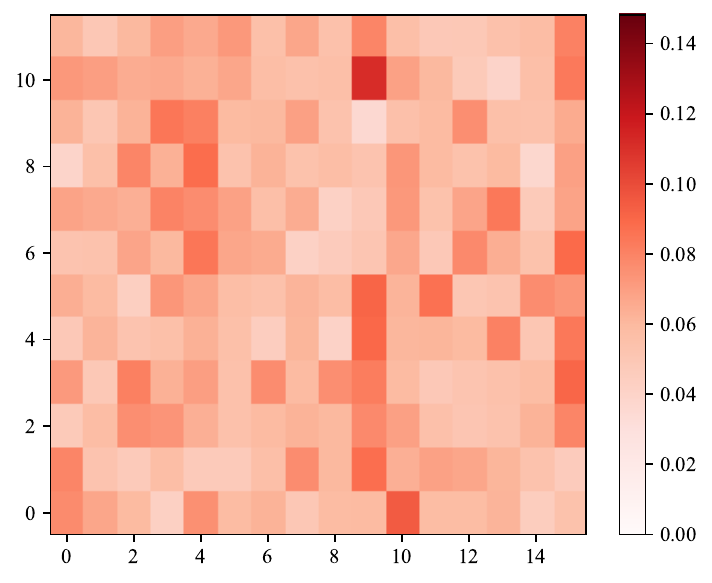}
		\centering
		\text{( \textit{b} )} 
	\end{minipage}
	\hfill
	\begin{minipage}[b]{0.24\textwidth}
		\includegraphics[width=\textwidth]{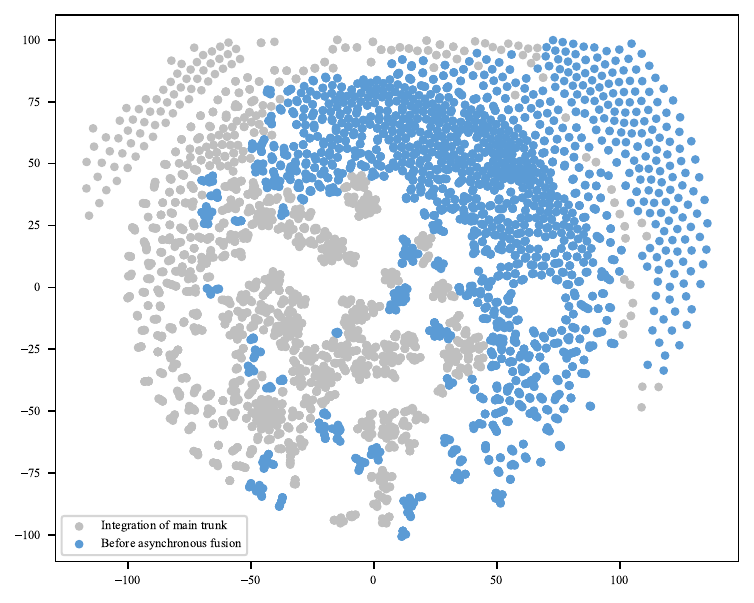}
		\centering
		\text{( \textit{c} )}
	\end{minipage}
	\hfill
	\begin{minipage}[b]{0.24\textwidth}
		\includegraphics[width=\textwidth]{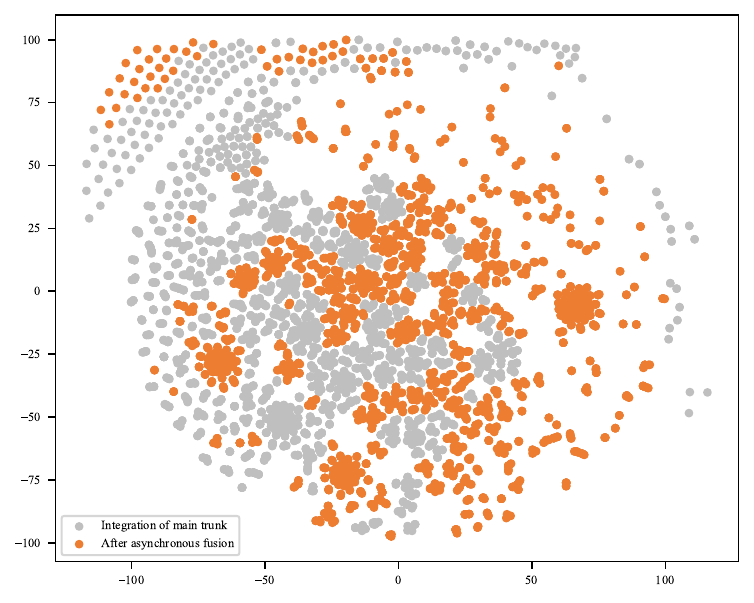}
		\centering
		\text{( \textit{d} )}
	\end{minipage}
	\caption{Visualization of the proposed asynchronous fusion mechanism on the Exchange dataset. (a) Cross-attention maps from fusion queries to temporal patches in the fusion stage. (b) Cross-attention maps from temporal patches to fusion queries in the refinement stage. (c) t-SNE projection of temporal features and fusion features before refinement. (d) t-SNE projection of temporal features and fusion features after refinement.}
	\label{fig:at}
\end{figure*}

\subsection{Ablation Study}
To rigorously assess the contribution of each component in TimeSAF, we conduct ablation studies guided by four questions: \ding{202} Is the semantic fusion trunk necessary? \ding{203} Does explicitly modeling fusion query slots outperform directly aggregating on unimodal features? \ding{204} Does gated semantic injection help stabilize asynchronous refinement? \ding{205} Is stage-wise asynchronous interaction superior to layer-wise synchronous updates?

Accordingly, we construct four variants: \textbf{w/o Fusion Trunk}, which removes the semantic fusion trunk;\textbf{ w/o Fusion Query}, which discards the independent fusion query slots and performs cross-modal interaction directly on unimodal backbone features; \textbf{w/o Gate}, which removes the scalar gating factor; and \textbf{Synchronous Refinement}, which replaces the proposed stage-wise asynchronous scheme with synchronous refinement within each layer. As shown in Fig.~\ref{fig:ab}, all variants suffer noticeable performance drops compared to the full TimeSAF model, confirming the utility of each component. Notably, Synchronous Refinement performs worst on most datasets, as synchronous fusion fails to mitigate semantic–perceptual dissonance, preventing high-level semantics from being properly formed and from effectively guiding low-level temporal representations. Overall, these results validate both the necessity of the proposed modules and the effectiveness of the hierarchical asynchronous fusion architecture.

\subsection{Attention Flow Visualization}
To qualitatively examine how the proposed hierarchical asynchronous fusion operates, we visualize the cross-attention maps between the semantic fusion trunk and the temporal trunk. As shown in Fig.~\ref{fig:at}(a)–(b), we plot (i) the attention from fusion queries to temporal patches at the fusion stage, and (ii) the attention from temporal patches back to fusion queries at the subsequent refinement stage. The two maps exhibit highly consistent patterns: temporal regions that receive strong attention from certain fusion queries during fusion tend to query the same slots during refinement. In other words, the time positions selected and aggregated by the fusion trunk later become the main recipients of its top-down feedback. This aligned attention flow supports our design intuition that the model first uses the fusion trunk to gather information from salient temporal regions, and then reuses the fused semantics to refine exactly those regions, forming a stable “aggregate–refine” loop that is absent in conventional synchronous fusion schemes.

\subsection{T-SNE Visualization}
We further examine the effect of asynchronous refinement from the perspective of the representation space. For a fixed refinement stage, we project (i) temporal features before refinement, (ii) temporal features after refinement, and (iii) the corresponding fusion features into a shared two-dimensional space using t-SNE, as shown in Fig.~\ref{fig:at}(c)–(d). Before refinement, the temporal features form a cluster clearly separated from the fusion features, indicating a mismatch between unimodal temporal representations and the fused semantic space. After applying the asynchronous refinement module, the temporal cluster moves toward and partially overlaps with the fusion cluster, suggesting that the refined temporal representations become better aligned with the fusion space. This embedding-level alignment supports that the proposed asynchronous fusion mechanism effectively narrows the representational gap between the temporal backbone and the fusion backbone.

\begin{figure}[!ht]  
	\centering
	\begin{minipage}[b]{0.5\textwidth}
		\includegraphics[width=\textwidth]{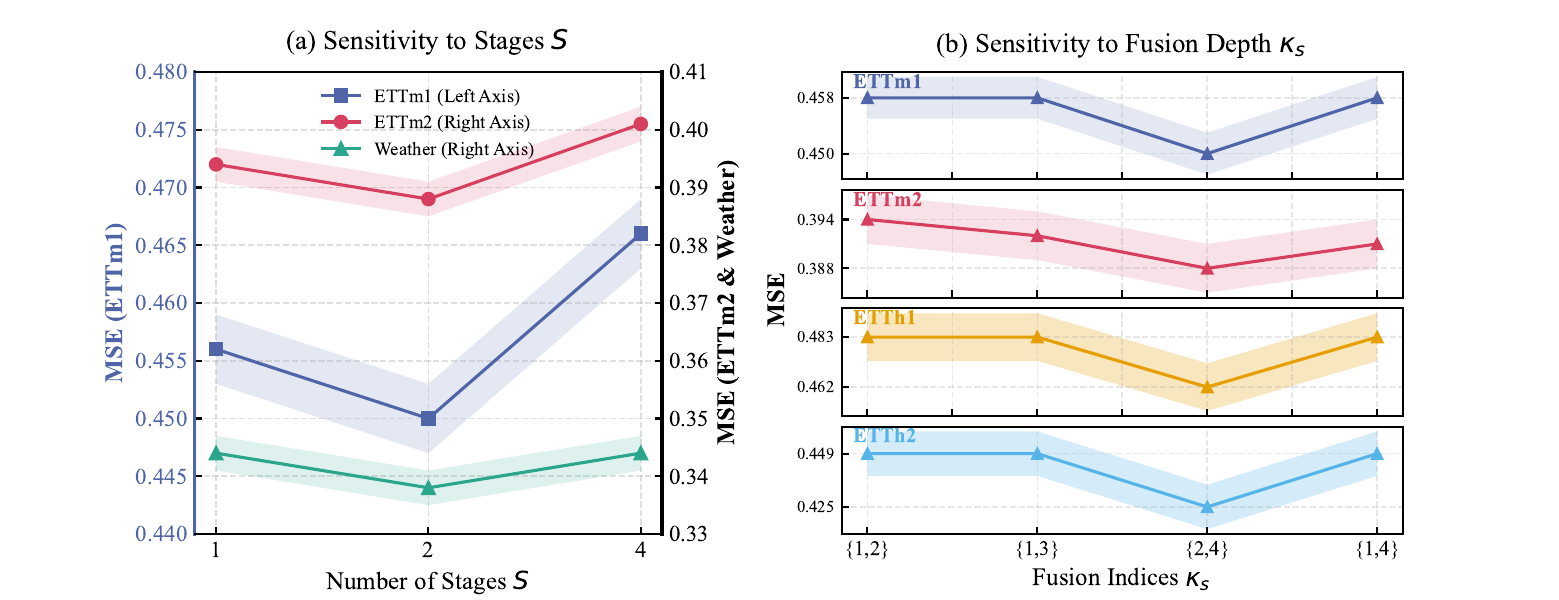}
		\centering
	\end{minipage}
	\caption{Sensitivity of TimeSAF to fusion configuration.}
	\label{fig:canshu}
\end{figure}
\subsection{Sensitivity to stage-wise fusion}
As shown in Fig. \ref{fig:canshu}. We first vary the number of fusion stages $S \in \{1,2,4\}$ on several datasets. 
Moving from a single stage to $S=2$ consistently reduces the prediction error, while further increasing to $S=4$ brings little or no benefit, indicating that a small number of fusion stages is sufficient. 
Fixing $S=2$, we then shift the fusion layer indices $\kappa_s$ and observe that placing fusion blocks at middle/deeper layers yields slightly better results than very shallow or edge-heavy placements. 
Overall, TimeSAF is reasonably robust to $S$ and $\kappa_s$, and we adopt $S=2$ with a middle–deep fusion placement as the default setting.

\section{Conclusion}
We presented TimeSAF, a hierarchical asynchronous fusion framework for multimodal time-series forecasting. By decoupling unimodal encoding from cross-modal interaction and introducing a semantic fusion trunk with stage-wise refinement, TimeSAF mitigates semantic perceptual dissonance and allows LLM priors to guide temporal modeling more reliably. Experiments on standard LTSF benchmarks, as well as few-shot and zero-shot settings, show that TimeSAF achieves competitive or superior performance over strong Transformer-based and LLM-enhanced baselines, while remaining conceptually simple and practically deployable.

\section*{Limitations}
This work still has several limitations. First, TimeSAF is mainly evaluated on a set of standard LTSF benchmarks and constructed few-shot/zero-shot settings, and lacks large-scale studies in real industrial deployments or more complex multimodal environments. Our current implementation relies on rule-based templates to convert time-series statistics into natural language prompts. While this design is effective, it may not fully exploit the reasoning capability of language models compared with leveraging rich unstructured external knowledge (e.g., financial news or weather reports). Due to hardware constraints, our experiments primarily use GPT-2 as the semantic backbone. Although TimeSAF itself is model-agnostic, we have not yet systematically explored its behavior when scaled to larger or more advanced foundation models (such as LLaMA-3 or GPT-4). Future work can conduct more extensive empirical validation on larger-scale and more diverse real-world tasks.

\section*{Ethical considerations}
This work studies {TimeSAF} on public, aggregated benchmark datasets (ETT, Electricity, Weather, Exchange), which do not contain identifiable personal information. We adhere to the licenses and usage policies of the original data providers and do not introduce any additional sensitive data. Although TimeSAF is a generic forecasting framework, we do not conduct a dedicated fairness or bias analysis, and applying the model in high-stakes domains should therefore involve domain experts and proper risk assessment. Our experiments use a frozen medium-scale GPT-2 backbone and moderate GPU resources, which limits but does not eliminate the environmental footprint. We encourage responsible use of TimeSAF as a decision-support tool rather than an autonomous decision-maker, and discourage applications that lack transparency or may cause societal harm.

\section*{Acknowledgements}
This work was supported in part by the following: the National Natural Science Foundation of China under Grant Nos. U24A20219, 62272281, U24A20328, 62576193, the Yantai Natural Science Foundation under Grant No. 2024JCYJ034, and the Youth Innovation Technology Project of Higher School in Shandong Province under Grant No. 2023KJ212.

\bibliography{custom}

\appendix

\begin{table*}[!ht]   
	\caption{All experiments fix the lookback length \textit{T} = 96. The prediction length set is \textit{H} $ \in $ \{96, 192, 336, 720\}. The best result is \textcolor{red}{\textbf{red}}, the second best result is \textcolor{blue}{\underline{underlined}}. }
	\label{tab:allmsefull}
	\centering
	\renewcommand{\arraystretch}{1.2}
	\tiny 
	\setlength{\tabcolsep}{1.9pt} 
	\begin{tabular}{cccc|cc|cc|cc|cc|cc|cc|cc|cc|cc|cc}
		\toprule[1.2pt]
		
		\multirow{1}{*}{\textbf{Models}} & \multirow{2}{*}{\textbf{}} & \multicolumn{2}{c|}{\textbf{TimeSAF}} & \multicolumn{2}{c|}{\textbf{CALF}} & \multicolumn{2}{c|}{\textbf{TimeCMA}}& \multicolumn{2}{c|}{\textbf{Time-FFM}} & \multicolumn{2}{c|}{\textbf{UniTime}} & \multicolumn{2}{c}{\textbf{Time-LLM}}  & \multicolumn{2}{c|}{\textbf{GPT4TS}} & \multicolumn{2}{c|}{\textbf{iTransformer}} & \multicolumn{2}{c|}{\textbf{PatchTST}} & \multicolumn{2}{c|}{\textbf{Crossformer}} & \multicolumn{2}{c}{\textbf{FEDformer}} \\

		\cmidrule(r){1-24}
		\multirow{1}{*}{\textbf{Metrics}} 
		&{\textbf{}} & \textbf{MSE} & \multicolumn{1}{c|}{\textbf{MAE}} & \textbf{MSE}  & \multicolumn{1}{c|}{\textbf{MAE}}& \textbf{MSE} & \multicolumn{1}{c|}{\textbf{MAE}}& \textbf{MSE}  & \multicolumn{1}{c|}{\textbf{MAE}} & \textbf{MSE}  & \multicolumn{1}{c|}{\textbf{MAE}} & \textbf{MSE} & \multicolumn{1}{c|}{\textbf{MAE}} & \textbf{MSE}  & \multicolumn{1}{c|}{\textbf{MAE}} & \textbf{MSE} & \multicolumn{1}{c|}{\textbf{MAE}}& \textbf{MSE}  & \multicolumn{1}{c|}{\textbf{MAE}} & \textbf{MSE}  & \multicolumn{1}{c|}{\textbf{MAE}}&\textbf{MSE} & \multicolumn{1}{c}{\textbf{MAE}}\\
		\midrule
		\multirow{5}{*}{\textbf{ETTm1}} 
	& 96 & \textbf{\textcolor{red}{0.311}} &  \textcolor{blue}{\underline{0.351}} &  \textcolor{blue}{\underline{0.319}} & \textbf{\textcolor{red}{0.348}} & 0.330 & 0.371 & 0.336 & 0.369 & 0.322 & 0.363 & 0.359 & 0.381 & 0.335 & 0.369 
	& 0.334 & 0.368 & 0.329 & 0.367 & 0.360 & 0.401 & 0.379 & 0.419 \\
	& 192 & \textbf{\textcolor{red}{0.357}} & \textbf{\textcolor{red}{0.372}} & 0.373 &  \textcolor{blue}{\underline{0.375}} & 0.368 & 0.389 & 0.378 & 0.389 &  \textcolor{blue}{\underline{0.366}} & 0.387 & 0.383 & 0.393 & 0.374 & 0.385 & 0.377 & 0.391 & 0.367 & 0.385 & 0.402 & 0.440 & 0.426 & 0.441 \\
	& 336 & \textbf{\textcolor{red}{0.390}} & \textbf{\textcolor{red}{0.399}} & 0.411 &  \textcolor{blue}{\underline{0.400}} & 0.402 & 0.411 & 0.411 & 0.410 &  \textcolor{blue}{\underline{0.398}} & 0.407 & 0.416 & 0.414 & 0.407 & 0.406 & 0.426 & 0.420 & 0.399 & 0.410 & 0.543 & 0.528 & 0.445 & 0.459 \\
	& 720 & \textbf{\textcolor{red}{0.450}} & \textbf{\textcolor{red}{0.438}} & 0.478 &  \textcolor{blue}{\underline{0.438}} & 0.471 & 0.450 & 0.469 & 0.441 &  \textcolor{blue}{\underline{0.454}} & 0.440 & 0.483 & 0.449 & 0.469 & 0.442 & 0.491 & 0.459 & 0.454 & 0.439 & 0.704 & 0.642 & 0.543 & 0.490 \\
	& avg & \textbf{\textcolor{red}{0.377}} & \textbf{\textcolor{red}{0.390}} & 0.395 &  \textcolor{blue}{\underline{0.390}} & 0.392 & 0.405 & 0.399 & 0.402 &  \textcolor{blue}{\underline{0.385}} & 0.399 & 0.410 & 0.409 & 0.396 & 0.400 & 0.407 & 0.410 & 0.387 & 0.400 & 0.502 & 0.502 & 0.448 & 0.452 \\
		\midrule
		\multirow{5}{*}{\textbf{ETTm2}} 
& 96 & \textbf{\textcolor{red}{0.171}} &  \textcolor{blue}{\underline{0.255}} &  \textcolor{blue}{\underline{0.175}} & \textbf{\textcolor{red}{0.253}} & 0.181 & 0.263 & 0.181 & 0.267 & 0.183 & 0.266 & 0.193 & 0.280 & 0.190 & 0.275 
& 0.180 & 0.264 & 0.175 & 0.259 & 0.273 & 0.356 & 0.203 & 0.287 \\
& 192 & \textbf{\textcolor{red}{0.236}} & \textbf{\textcolor{red}{0.299}} & 0.242 &  \textcolor{blue}{\underline{0.299}} & 0.256 & 0.315 & 0.247 & 0.308 & 0.251 & 0.310 & 0.257 & 0.318 & 0.253 & 0.313 & 0.250 & 0.309 &  \textcolor{blue}{\underline{0.241}} & 0.302 & 0.426 & 0.487 & 0.269 & 0.328 \\
& 336 & \textbf{\textcolor{red}{0.294}} & \textbf{\textcolor{red}{0.336}} &  \textcolor{blue}{\underline{0.303}} &  \textcolor{blue}{\underline{0.337}} & 0.309 & 0.347 & 0.309 & 0.347 & 0.319 & 0.351 & 0.317 & 0.353 & 0.321 & 0.360 & 0.311 & 0.348 & 0.305 & 0.343 & 1.013 & 0.714 & 0.325 & 0.366 \\
& 720 & \textbf{\textcolor{red}{0.388}} & \textbf{\textcolor{red}{0.393}} & 0.409 &  \textcolor{blue}{\underline{0.398}} & 0.411 & 0.404 & 0.406 & 0.404 & 0.420 & 0.410 & 0.419 & 0.411 & 0.411 & 0.406 & 0.412 & 0.407 &  \textcolor{blue}{\underline{0.402}} & 0.400 & 3.154 & 1.274 & 0.421 & 0.415 \\
& avg & \textbf{\textcolor{red}{0.272}} & \textbf{\textcolor{red}{0.320}} & 0.282 &  \textcolor{blue}{\underline{0.321}} & 0.289 & 0.332 & 0.286 & 0.332 & 0.293 & 0.334 & 0.296 & 0.340 & 0.293 & 0.338 & 0.288 & 0.332 &  \textcolor{blue}{\underline{0.281}} & 0.326 & 1.216 & 0.707 & 0.304 & 0.349 \\
		\midrule
		\multirow{5}{*}{\textbf{ETTh1}} 
		
	& 96 & \textbf{\textcolor{red}{0.369}} & \textbf{\textcolor{red}{0.391}} &  \textcolor{blue}{\underline{0.376}} &  \textcolor{blue}{\underline{0.393}} & 0.399 & 0.402 & 0.385 & 0.400 & 0.397 & 0.418 & 0.398 & 0.410 & 0.398 & 0.424 
	& 0.386 & 0.405 & 0.414 & 0.419 & 0.420 & 0.439 & 0.376 & 0.419 \\
	& 192 &  \textcolor{blue}{\underline{0.423}} & \textbf{\textcolor{red}{0.420}} & 0.432 &  \textcolor{blue}{\underline{0.427}} & 0.442 & 0.441 & 0.439 & 0.430 & 0.434 & 0.439 & 0.451 & 0.440 & 0.449 & 0.427 & 0.441 & 0.436 & 0.460 & 0.445 & 0.540 & 0.519 & \textbf{\textcolor{red}{0.420}} & 0.448 \\
	& 336 & \textbf{\textcolor{red}{0.451}} & \textbf{\textcolor{red}{0.438}} & 0.479 & 0.451 & 0.477 & 0.477 & 0.480 &  \textcolor{blue}{\underline{0.449}} & 0.468 & 0.457 & 0.508 & 0.471 & 0.492 & 0.466 & 0.487 & 0.458 & 0.501 & 0.466 & 0.722 & 0.648 &  \textcolor{blue}{\underline{0.459}} & 0.465 \\
	& 720 & \textbf{\textcolor{red}{0.462}} &  \textcolor{blue}{\underline{0.466}} & 0.486 & 0.472 & 0.488 & 0.478 &  \textcolor{blue}{\underline{0.462}} & \textbf{\textcolor{red}{0.456}} & 0.469 & 0.477 & 0.483 & 0.478 & 0.487 & 0.483 & 0.503 & 0.491 & 0.500 & 0.488 & 0.799 & 0.685 & 0.506 & 0.507 \\
	& avg & \textbf{\textcolor{red}{0.426}} & \textbf{\textcolor{red}{0.428}} & 0.443 & 0.435 & 0.451 & 0.449 & 0.442 &  \textcolor{blue}{\underline{0.434}} & 0.442 & 0.447 & 0.460 & 0.449 & 0.456 & 0.450 & 0.454 & 0.447 & 0.469 & 0.454 & 0.620 & 0.572 &  \textcolor{blue}{\underline{0.440}} & 0.459 \\
		\midrule
		\multirow{5}{*}{\textbf{ETTh2}} 
	& 96 & \textbf{\textcolor{red}{0.286}} &  \textcolor{blue}{\underline{0.338}} &  \textcolor{blue}{\underline{0.290}} & \textbf{\textcolor{red}{0.337}} & 0.326 & 0.363 & 0.301 & 0.351 & 0.296 & 0.345 & 0.295 & 0.346 & 0.398 & 0.360 
	& 0.297 & 0.349 & 0.302 & 0.348 & 0.745 & 0.584 & 0.358 & 0.397 \\
	& 192 &  \textcolor{blue}{\underline{0.370}} &  \textcolor{blue}{\underline{0.390}} & \textbf{\textcolor{red}{0.366}} & \textbf{\textcolor{red}{0.385}} & 0.420 & 0.420 & 0.378 & 0.397 & 0.374 & 0.394 & 0.386 & 0.399 & 0.449 & 0.405 & 0.380 & 0.400 & 0.388 & 0.400 & 0.877 & 0.656 & 0.429 & 0.439 \\
	& 336 & \textbf{\textcolor{red}{0.391}} & \textbf{\textcolor{red}{0.410}} & 0.416 &  \textcolor{blue}{\underline{0.423}} & 0.442 & 0.444 & 0.422 & 0.431 &  \textcolor{blue}{\underline{0.415}} & 0.427 & 0.447 & 0.443 & 0.492 & 0.437 & 0.428 & 0.432 & 0.426 & 0.433 & 1.043 & 0.731 & 0.496 & 0.487 \\
	& 720 &  \textcolor{blue}{\underline{0.425}} &  \textcolor{blue}{\underline{0.438}} & \textbf{\textcolor{red}{0.415}} & \textbf{\textcolor{red}{0.434}} & 0.442 & 0.452 & 0.427 & 0.444 & 0.425 & 0.444 & 0.428 & 0.444 & 0.487 & 0.453 & 0.427 & 0.445 & 0.431 & 0.446 & 1.104 & 0.763 & 0.463 & 0.474 \\
	& avg & \textbf{\textcolor{red}{0.367}} & \textbf{\textcolor{red}{0.394}} &  \textcolor{blue}{\underline{0.371}} &  \textcolor{blue}{\underline{0.394}} & 0.407 & 0.419 & 0.382 & 0.406 & 0.377 & 0.402 & 0.389 & 0.408 & 0.456 & 0.413 & 0.383 & 0.407 & 0.387 & 0.407 & 0.942 & 0.683 & 0.436 & 0.449 \\
		\midrule
		\multirow{5}{*}{\textbf{Weather
		}} 
& 96 & \textbf{\textcolor{red}{0.157}} & \textbf{\textcolor{red}{0.200}} & 0.166 &  \textcolor{blue}{\underline{0.205}} & 0.176 & 0.221 & 0.191 & 0.230 & 0.171 & 0.214 & 0.195 & 0.233 & 0.203 & 0.244 & 0.174 & 0.214 & 0.177 
& 0.218 &  \textcolor{blue}{\underline{0.158}} & 0.230 & 0.217 & 0.296 \\
& 192 & \textbf{\textcolor{red}{0.204}} & \textbf{\textcolor{red}{0.247}} & 0.214 &  \textcolor{blue}{\underline{0.253}} & 0.215 & 0.256 & 0.236 & 0.267 & 0.217 & 0.254 & 0.240 & 0.269 & 0.247 & 0.277 & 0.221 & 0.254 & 0.225 & 0.259 &  \textcolor{blue}{\underline{0.206}} & 0.277 & 0.276 & 0.336 \\
& 336 & \textbf{\textcolor{red}{0.262}} & \textbf{\textcolor{red}{0.287}} &  \textcolor{blue}{\underline{0.268}} &  \textcolor{blue}{\underline{0.291}} & 0.276 & 0.302 & 0.289 & 0.303 & 0.274 & 0.293 & 0.293 & 0.306 & 0.297 & 0.311 & 0.278 & 0.296 & 0.278 & 0.297 & 0.272 & 0.335 & 0.339 & 0.380 \\
& 720 & \textbf{\textcolor{red}{0.338}} & \textbf{\textcolor{red}{0.340}} &  \textcolor{blue}{\underline{0.348}} & 0.345 & 0.348 & 0.353 & 0.362 & 0.350 & 0.351 &  \textcolor{blue}{\underline{0.343}} & 0.368 & 0.354 & 0.368 & 0.356 & 0.358 & 0.347 & 0.354 & 0.348 & 0.398 & 0.418 & 0.403 & 0.428 \\
& avg & \textbf{\textcolor{red}{0.240}} & \textbf{\textcolor{red}{0.268}} &  \textcolor{blue}{\underline{0.249}} &  \textcolor{blue}{\underline{0.273}} & 0.253 & 0.283 & 0.270 & 0.288 & 0.253 & 0.276 & 0.274 & 0.290 & 0.278 & 0.297 & 0.257 & 0.278 & 0.259 & 0.281 & 0.258 & 0.315 & 0.308 & 0.360 \\
		\midrule
		\multirow{5}{*}{\textbf{Exchange}} 
& 96 & \textbf{\textcolor{red}{0.078}} & \textbf{\textcolor{red}{0.196}} & 0.082 &  \textcolor{blue}{\underline{0.201}} & 0.097 & 0.221 &  \textcolor{blue}{\underline{0.081}} & 0.201 & 0.086 & 0.209 & 0.087 & 0.208 & 0.091 & 0.212 
& 0.086 & 0.206 & 0.088 & 0.205 & 0.139 & 0.265 & 0.148 & 0.278 \\
& 192 &  \textcolor{blue}{\underline{0.170}} & \textbf{\textcolor{red}{0.293}} & 0.173 & 0.296 & 0.195 & 0.320 & \textbf{\textcolor{red}{0.168}} &  \textcolor{blue}{\underline{0.293}} & 0.174 & 0.299 & 0.173 & 0.299 & 0.183 & 0.304 & 0.177 & 0.299 & 0.176 & 0.299 & 0.241 & 0.375 & 0.271 & 0.315 \\
& 336 & 0.312 & 0.406 & 0.345 & 0.425 & 0.372 & 0.449 & \textbf{\textcolor{red}{0.299}} & \textbf{\textcolor{red}{0.396}} & 0.319 & 0.408 & 0.375 & 0.454 & 0.328 & 0.417 & 0.331 & 0.417 &  \textcolor{blue}{\underline{0.301}} &  \textcolor{blue}{\underline{0.397}} & 0.392 & 0.468 & 0.460 & 0.427 \\
& 720 & \textbf{\textcolor{red}{0.801}} & \textbf{\textcolor{red}{0.647}} & 0.824 & 0.686 & 1.067 & 0.794 &  \textcolor{blue}{\underline{0.805}} &  \textcolor{blue}{\underline{0.674}} & 0.875 & 0.701 & 0.853 & 0.703 & 0.880 & 0.704 & 0.847 & 0.691 & 0.901 & 0.714 & 1.110 & 0.802 & 1.195 & 0.695 \\
& avg &  \textcolor{blue}{\underline{0.340}} & \textbf{\textcolor{red}{0.385}} & 0.356 & 0.402 & 0.432 & 0.446 & \textbf{\textcolor{red}{0.338}} &  \textcolor{blue}{\underline{0.391}} & 0.363 & 0.404 & 0.372 & 0.416 & 0.370 & 0.409 & 0.360 & 0.403 & 0.367 & 0.404 & 0.470 & 0.477 & 0.518 & 0.428 \\

		\midrule
		\multirow{5}{*}{\textbf{Electricity}} 
& 96 & \textbf{\textcolor{red}{0.143}} & \textbf{\textcolor{red}{0.237}} &  \textcolor{blue}{\underline{0.145}} &  \textcolor{blue}{\underline{0.239}} & 0.146 & 0.247 & 0.198 & 0.282 & 0.196 & 0.287 & 0.204 & 0.293 & 0.197 & 0.290 
& 0.148 & 0.240 & 0.181 & 0.270 & 0.219 & 0.314 & 0.193 & 0.308 \\
& 192 & \textbf{\textcolor{red}{0.160}} & 0.254 &  \textcolor{blue}{\underline{0.161}} & \textbf{\textcolor{red}{0.252}} & 0.163 & 0.264 & 0.199 & 0.285 & 0.199 & 0.291 & 0.207 & 0.295 & 0.201 & 0.292 & 0.162 &  \textcolor{blue}{\underline{0.253}} & 0.188 & 0.274 & 0.231 & 0.322 & 0.201 & 0.315 \\
& 336 & \textbf{\textcolor{red}{0.176}} & \textbf{\textcolor{red}{0.266}} &  \textcolor{blue}{\underline{0.177}} &  \textcolor{blue}{\underline{0.269}} & 0.223 & 0.323 & 0.212 & 0.298 & 0.214 & 0.305 & 0.219 & 0.308 & 0.217 & 0.309 & 0.178 & 0.269 & 0.204 & 0.293 & 0.246 & 0.337 & 0.214 & 0.329 \\
& 720 & \textbf{\textcolor{red}{0.219}} & \textbf{\textcolor{red}{0.300}} &  \textcolor{blue}{\underline{0.219}} &  \textcolor{blue}{\underline{0.302}} & 0.245 & 0.317 & 0.253 & 0.330 & 0.254 & 0.335 & 0.263 & 0.341 & 0.253 & 0.339 & 0.225 & 0.317 & 0.246 & 0.324 & 0.280 & 0.363 & 0.246 & 0.355 \\
& avg & \textbf{\textcolor{red}{0.174}} & \textbf{\textcolor{red}{0.264}} &  \textcolor{blue}{\underline{0.175}} &  \textcolor{blue}{\underline{0.265}} & 0.194 & 0.287 & 0.216 & 0.299 & 0.215 & 0.304 & 0.223 & 0.309 & 0.217 & 0.307 & 0.178 & 0.270 & 0.205 & 0.290 & 0.244 & 0.334 & 0.213 & 0.326 \\	
		
		\bottomrule[1.2pt]
	\end{tabular}
	\vspace{0.5em}
\end{table*}

\section{Performance of Long-term Multivariate Forecasting}\label{sec:full}
This appendix reports the complete numerical results for all forecasting experiments.  
Table~\ref{tab:allmsefull} summarizes long-term multivariate forecasting on all benchmarks, using a fixed input length of $L = 96$ and four prediction horizons $H \in \{96, 192, 336, 720\}$.  
Table ~\ref{tab:10full} provides the corresponding few-shot results, where each model uses only 10\% of the original training set, $L = 96$ and $H \in \{96, 192, 336, 720\}$.
Table~\ref{tab:0full} reports the zero-shot transfer setting, in which models are trained on a source dataset and directly evaluated on a different target dataset without any further fine-tuning, still with $L = 96$ and $H \in \{96, 192, 336, 720\}$.  For all tables, we list MSE and MAE for TimeSAF and all baselines.
\section{Dataset Descriptions}\label{app:datasets}
We extensively evaluate our model on seven widely recognized real-world datasets, covering diverse domains such as energy, weather, and economics. Table \ref{tab:dataset_stat} summarizes the key statistics of these datasets. Specifically, the ETT dataset is divided into training, validation, and test sets with a ratio of 6:2:2, whereas all other datasets follow a 7:1:2 split. 

\begin{itemize}
	\item \textbf{ETT (Electricity Transformer Temperature):} This dataset comprises two years of data (July 2016 to July 2018) collected from electricity transformers, including oil temperature and power load features. It is categorized into four subsets based on sampling frequency: \textbf{ETTh1/ETTh2} (hourly) and \textbf{ETTm1/ETTm2} (every 15 minutes), allowing for evaluation at different temporal granularities.
	
	\item \textbf{Electricity:} This dataset monitors the hourly electricity consumption (in kW) of 321 clients from 2012 to 2014. The timestamps follow Portuguese time, requiring specific handling for Daylight Saving Time (DST). Specifically, measurements during the skipped hour in March (1:00 AM - 2:00 AM) are set to zero, while the overlapping hour in October is handled by aggregating the values.

	\item \textbf{Weather:} Recorded by the Max Planck Institute for Biogeochemistry, this dataset consists of 21 meteorological indicators collected every 10 minutes throughout 2020, capturing fine-grained climatic variations.
	
	\item \textbf{Exchange:} This financial dataset tracks the daily exchange rates of eight major countries (Australia, UK, Canada, Switzerland, China, Japan, New Zealand, and Singapore) against the US dollar, covering a long historical period from 1990 to 2016.
\end{itemize}

\begin{table*}[!ht]   
	\caption{Few-shot forecasting performance on ETT datasets using only 10\% of the training data. The prediction horizon is set to \textit{H} \( \in \) \{96, 192, 336, 720\}. We report the average MSE and MAE over all horizons; The best result is \textcolor{red}{\textbf{red}}, the second best result is \textcolor{blue}{\underline{underlined}}.}
	\label{tab:10full}
	\centering
	\renewcommand{\arraystretch}{1.3}
	\tiny 
	\setlength{\tabcolsep}{2.5pt} 
	\begin{tabular}{cccc|cc|cc|cc|cc|cc|cc|cc|cc|cc|cc}
		\toprule[1.2pt]
		
		\multirow{1}{*}{\textbf{Models}} & \multirow{2}{*}{\textbf{}} & \multicolumn{2}{c|}{\textbf{TimeSAF}} & \multicolumn{2}{c|}{\textbf{CALF\textsuperscript{\textdagger}}} & \multicolumn{2}{c|}{\textbf{TimeCMA\textsuperscript{\textdagger}}} & \multicolumn{2}{c|}{\textbf{Time-LLM }} & \multicolumn{2}{c|}{\textbf{GPT4TS}}  & \multicolumn{2}{c|}{\textbf{PatchTST}} & \multicolumn{2}{c|}{\textbf{Crossformer}} & \multicolumn{2}{c|}{\textbf{FEDformer}} & \multicolumn{2}{c|}{\textbf{TimesNet}}  & \multicolumn{2}{c|}{\textbf{DLinear}} & \multicolumn{2}{c}{\textbf{MICN}} \\

		\cmidrule(r){1-24}
		\multirow{1}{*}{\textbf{Metrics}} 
		&{\textbf{}} & \textbf{MSE} & \multicolumn{1}{c|}{\textbf{MAE}} & \textbf{MSE}  & \multicolumn{1}{c|}{\textbf{MAE}}& \textbf{MSE} & \multicolumn{1}{c|}{\textbf{MAE}}& \textbf{MSE}  & \multicolumn{1}{c|}{\textbf{MAE}} & \textbf{MSE}  & \multicolumn{1}{c|}{\textbf{MAE}} & \textbf{MSE} & \multicolumn{1}{c|}{\textbf{MAE}} & \textbf{MSE}  & \multicolumn{1}{c|}{\textbf{MAE}} & \textbf{MSE} & \multicolumn{1}{c|}{\textbf{MAE}}& \textbf{MSE}  & \multicolumn{1}{c|}{\textbf{MAE}} & \textbf{MSE}  & \multicolumn{1}{c|}{\textbf{MAE}}&\textbf{MSE} & \multicolumn{1}{c}{\textbf{MAE}}\\
		\midrule
		\multirow{5}{*}{\textbf{ETTm1}} 
		& 96 & \textbf{\textcolor{red}{0.442}} & \textbf{\textcolor{red}{0.430}} &  \textcolor{blue}{\underline{0.468}} &  \textcolor{blue}{\underline{0.445}} & 0.568 & 0.483 & 0.587 & 0.491 & 0.615 & 0.497 & 0.558 & 0.478 & 1.037 & 0.705 & 0.604 & 0.530 & 0.583 & 0.503 & 0.552 & 0.488 & 0.677 & 0.585 \\
		& 192 & \textbf{\textcolor{red}{0.457}} & \textbf{\textcolor{red}{0.437}} &  \textcolor{blue}{\underline{0.479}} &  \textcolor{blue}{\underline{0.446}} & 0.532 & 0.475 & 0.606 & 0.490 & 0.597 & 0.492 & 0.539 & 0.471 & 1.170 & 0.778 & 0.641 & 0.546 & 0.608 & 0.515 & 0.546 & 0.487 & 0.784 & 0.627 \\
		& 336 & \textbf{\textcolor{red}{0.476}} & \textbf{\textcolor{red}{0.450}} &  \textcolor{blue}{\underline{0.499}} &  \textcolor{blue}{\underline{0.463}} & 0.548 & 0.489 & 0.719 & 0.555 & 0.597 & 0.501 & 0.558 & 0.488 & 1.463 & 0.913 & 0.768 & 0.606 & 0.733 & 0.572 & 0.567 & 0.501 & 0.972 & 0.684 \\
		& 720 & \textbf{\textcolor{red}{0.559}} & \textbf{\textcolor{red}{0.479}} &  \textcolor{blue}{\underline{0.572}} &  \textcolor{blue}{\underline{0.496}} & 0.729 & 0.571 & 0.632 & 0.514 & 0.623 & 0.513 & 0.574 & 0.498 & 1.693 & 0.997 & 0.771 & 0.606 & 0.768 & 0.548 & 0.606 & 0.522 & 1.449 & 0.800 \\
		& avg & \textbf{\textcolor{red}{0.483}} & \textbf{\textcolor{red}{0.449}} &  \textcolor{blue}{\underline{0.504}} &  \textcolor{blue}{\underline{0.462}} & 0.594 & 0.504 & 0.636 & 0.512 & 0.608 & 0.500 & 0.557 & 0.483 & 1.340 & 0.848 & 0.696 & 0.572 & 0.673 & 0.534 & 0.567 & 0.499 & 0.970 & 0.674 \\
		\midrule
		\multirow{5}{*}{\textbf{ETTm2}} 
		& 96 & \textbf{\textcolor{red}{0.182}} & \textbf{\textcolor{red}{0.261}} & 0.190 & 0.268 & 0.205 & 0.280 & 0.189 & 0.270 &  \textcolor{blue}{\underline{0.187}} &  \textcolor{blue}{\underline{0.266}} & 0.189 & 0.268 & 1.397 & 0.866 & 0.222 & 0.314 & 0.214 & 0.288 & 0.225 & 0.320 & 0.389 & 0.448 \\
		& 192 & \textbf{\textcolor{red}{0.248}} & \textbf{\textcolor{red}{0.303}} & 0.257 & 0.311 & 0.269 & 0.320 & 0.264 & 0.319 & 0.253 & 0.308 &  \textcolor{blue}{\underline{0.248}} &  \textcolor{blue}{\underline{0.307}} & 1.757 & 0.987 & 0.284 & 0.351 & 0.271 & 0.325 & 0.291 & 0.362 & 0.622 & 0.575 \\
		& 336 & \textbf{\textcolor{red}{0.311}} &  \textcolor{blue}{\underline{0.343}} & 0.323 & \textbf{\textcolor{red}{0.334}} & 0.325 & 0.354 & 0.327 & 0.358 & 0.332 & 0.353 &  \textcolor{blue}{\underline{0.311}} & 0.346 & 2.075 & 1.086 & 0.392 & 0.419 & 0.329 & 0.356 & 0.354 & 0.402 & 1.055 & 0.755 \\
		& 720 & \textbf{\textcolor{red}{0.435}} &  \textcolor{blue}{\underline{0.416}} & 0.441 & \textbf{\textcolor{red}{0.410}} & 0.436 & 0.421 & 0.454 & 0.428 & 0.438 & 0.417 &  \textcolor{blue}{\underline{0.435}} & 0.418 & 2.712 & 1.253 & 0.527 & 0.485 & 0.473 & 0.448 & 0.446 & 0.447 & 2.226 & 1.087 \\
		& avg & \textbf{\textcolor{red}{0.294}} & \textbf{\textcolor{red}{0.330}} & 0.302 &  \textcolor{blue}{\underline{0.330}} & 0.309 & 0.344 & 0.308 & 0.343 & 0.303 & 0.336 &  \textcolor{blue}{\underline{0.295}} & 0.334 & 1.985 & 1.048 & 0.356 & 0.392 & 0.321 & 0.354 & 0.329 & 0.382 & 1.073 & 0.716 \\
		\midrule
		\multirow{5}{*}{\textbf{ETTh1}} 
		& 96 & \textbf{\textcolor{red}{0.430}} & \textbf{\textcolor{red}{0.426}} & 0.468 & 0.457 & 0.655 & 0.540 & 0.500 & 0.464 & 0.462 & 0.449 &  \textcolor{blue}{\underline{0.433}} &  \textcolor{blue}{\underline{0.428}} & 1.129 & 0.775 & 0.651 & 0.563 & 0.855 & 0.625 & 0.590 & 0.515 & 0.689 & 0.592 \\
		& 192 & \textbf{\textcolor{red}{0.491}} & \textbf{\textcolor{red}{0.459}} & 0.550 & 0.501 & 0.673 & 0.550 & 0.590 & 0.516 & 0.551 & 0.495 &  \textcolor{blue}{\underline{0.509}} &  \textcolor{blue}{\underline{0.474}} & 1.832 & 0.922 & 0.666 & 0.562 & 0.791 & 0.589 & 0.634 & 0.541 & 1.160 & 0.748 \\
		& 336 & 0.615 & 0.549 &  \textcolor{blue}{\underline{0.581}} &  \textcolor{blue}{\underline{0.521}} & 0.717 & 0.568 & 0.638 & 0.542 & 0.630 & 0.539 & \textbf{\textcolor{red}{0.572}} & \textbf{\textcolor{red}{0.509}} & 2.022 & 0.973 & 0.767 & 0.602 & 0.939 & 0.648 & 0.659 & 0.554 & 1.747 & 0.899 \\
		& 720 & 0.955 & 0.674 & 0.978 & 0.685 &  \textcolor{blue}{\underline{0.841}} &  \textcolor{blue}{\underline{0.641}} & 1.334 & 0.816 & 1.113 & 0.738 & 1.221 & 0.773 & 1.903 & 0.986 & 0.918 & 0.703 & 0.876 & 0.641 & \textbf{\textcolor{red}{0.708}} & \textbf{\textcolor{red}{0.598}} & 2.024 & 1.019 \\
		& avg & \textbf{\textcolor{red}{0.622}} & \textbf{\textcolor{red}{0.527}} &  \textcolor{blue}{\underline{0.644}} &  \textcolor{blue}{\underline{0.541}} & 0.721 & 0.575 & 0.765 & 0.584 & 0.689 & 0.555 & 0.683 & 0.645 & 1.744 & 0.914 & 0.750 & 0.607 & 0.865 & 0.625 & 0.647 & 0.552 & 1.405 & 0.814 \\
		\midrule
		\multirow{5}{*}{\textbf{ETTh2}} 
		& 96 & \textbf{\textcolor{red}{0.311}} & \textbf{\textcolor{red}{0.348}} &  \textcolor{blue}{\underline{0.314}} & 0.360 & 0.339 & 0.375 & 0.329 & 0.365 & 0.327 & 0.359 & 0.314 &  \textcolor{blue}{\underline{0.354}} & 2.482 & 1.206 & 0.359 & 0.404 & 0.372 & 0.405 & 
		0.361 & 0.407 & 0.510 & 0.502 \\
		& 192 & 0.412 & \textbf{\textcolor{red}{0.404}} &  \textcolor{blue}{\underline{0.404}} & 0.411 & 0.443 & 0.437 & 0.414 & 0.413 & \textbf{\textcolor{red}{0.403}} &  \textcolor{blue}{\underline{0.405}} & 0.420 & 0.415 & 3.136 & 1.372 & 0.460 & 0.461 & 0.483 & 0.463 & 0.444 & 0.453 & 1.809 & 1.036 \\
		& 336 & \textbf{\textcolor{red}{0.452}} & \textbf{\textcolor{red}{0.450}} &  \textcolor{blue}{\underline{0.458}} &  \textcolor{blue}{\underline{0.452}} & 0.513 & 0.486 & 0.579 & 0.506 & 0.568 & 0.499 & 0.543 & 0.489 & 2.925 & 1.331 & 0.569 & 0.530 & 0.541 & 0.496 & 0.509 & 0.501 & 3.250 & 1.419 \\
		& 720 & 0.513 & 0.493 &  \textcolor{blue}{\underline{0.502}} &  \textcolor{blue}{\underline{0.487}} & 0.506 & 0.492 & 1.034 & 0.711 & 1.020 & 0.725 & 0.926 & 0.691 & 4.014 & 1.603 & 0.827 & 0.707 & 0.510 & 0.491 & \textbf{\textcolor{red}{0.453}} & \textbf{\textcolor{red}{0.471}} & 4.564 & 1.676 \\
		& avg &  \textcolor{blue}{\underline{0.422}} & \textbf{\textcolor{red}{0.423}} & \textbf{\textcolor{red}{0.419}} &  \textcolor{blue}{\underline{0.427}} & 0.451 & 0.448 & 0.589 & 0.498 & 0.579 & 0.497 & 0.550 & 0.487 & 3.139 & 1.378 & 0.553 & 0.525 & 0.476 & 0.463 & 0.441 & 0.458 & 2.533 & 1.158 \\
		
		\bottomrule[1.2pt]
	\end{tabular}
	\vspace{0.5em}
\end{table*}
\begin{table*}[!ht]
	\caption{
		Full results for zero-shot forecasting on the ETT datasets, where prediction lengths $H \in \{96,192,336,720\}$.  “h1→m1” indicates that models trained on ETTh1 are evaluated on ETTm1, and the same applies to other items. The best result is \textcolor{red}{\textbf{red}}, the second best result is \textcolor{blue}{\underline{underlined}}. }
	\label{tab:0full}
	\centering
	\tiny
	\renewcommand{\arraystretch}{1.25}
	\setlength{\tabcolsep}{2.3pt}
	
	\begin{tabular}{c|c|cc|cc|cc|cc|cc|cc|cc|cc|cc|cc|cc}
		\toprule[1.2pt]
		\textbf{Models} & \textbf{Metric} &
		\multicolumn{2}{c|}{\textbf{TimeSAF}} &
		\multicolumn{2}{c|}{\textbf{CALF}} &
		\multicolumn{2}{c}{\textbf{TimeCMA}} &
		\multicolumn{2}{c|}{\textbf{Time-LLM}} &
		\multicolumn{2}{c|}{\textbf{GPT4TS }} &
		\multicolumn{2}{c|}{\textbf{PatchTST}} &
		\multicolumn{2}{c|}{\textbf{Crossformer}} &
		\multicolumn{2}{c|}{\textbf{FEDformer}} &
		\multicolumn{2}{c|}{\textbf{TimesNet}} &
		\multicolumn{2}{c|}{\textbf{MICN }} &
		\multicolumn{2}{c}{\textbf{DLinear}} \\
		\midrule
		
		\multirow{5}{*}{\textbf{h1 → m1}} 
		& 96 & \textbf{\textcolor{red}{0.729}} & \textbf{\textcolor{red}{0.544}} & 0.767 & 0.564 & 0.809 & 0.577 & 0.804 & 0.565 & 0.809 & 0.563 & 0.908 & 0.596 & 0.856 & 0.649 &  \textcolor{blue}{\underline{0.731}} & 0.561 & 0.764 & 0.563 & 0.832 & 0.621 & 0.735 & 
		\textcolor{blue}{\underline{0.554}} \\
		& 192 &  \textcolor{blue}{\underline{0.750}} & \textbf{\textcolor{red}{0.554}} & 0.753 & 0.570 & 0.814 & 0.584 & 0.827 & 0.593 & 0.799 & 0.567 & 0.927 & 0.616 & 0.906 & 0.684 & \textbf{\textcolor{red}{0.746}} & 0.573 & 0.798 &  \textcolor{blue}{\underline{0.562}} & 1.288 & 0.854 & 0.752 & 0.570 \\
		& 336 & \textbf{\textcolor{red}{0.744}} & \textbf{\textcolor{red}{0.564}} &  \textcolor{blue}{\underline{0.745}} &  \textcolor{blue}{\underline{0.575}} & 0.814 & 0.591 & 0.835 & 0.600 & 0.803 & 0.577 & 0.920 & 0.621 & 1.104 & 0.796 & 0.775 & 0.596 & 0.790 & 0.584 & 1.721 & 0.972 & 0.749 & 0.579 \\
		& 720 &  \textcolor{blue}{\underline{0.774}} & \textbf{\textcolor{red}{0.588}} & \textbf{\textcolor{red}{0.758}} & 0.590 & 0.841 & 0.607 & 0.922 & 0.644 & 0.783 &  \textcolor{blue}{\underline{0.589}} & 0.822 & 0.608 & 1.131 & 0.816 & 0.808 & 0.625 & 0.827 & 0.594 & 1.915 & 1.036 & 0.805 & 0.606 \\
		& avg & \textbf{\textcolor{red}{0.749}} & \textbf{\textcolor{red}{0.563}} &  \textcolor{blue}{\underline{0.755}} &  \textcolor{blue}{\underline{0.574}} & 0.820 & 0.590 & 0.847 & 0.600 & 0.798 & 0.574 & 0.894 & 0.610 & 0.999 & 0.736 & 0.765 & 0.588 & 0.794 & 0.575 & 1.439 & 0.870 & 0.760 & 0.577 \\	
		
		\midrule
		
		\multirow{5}{*}{\textbf{h1 → m2}} 
		& 96 &  \textcolor{blue}{\underline{0.217}} &  \textcolor{blue}{\underline{0.301}} & 0.218 & 0.301 & 0.229 & 0.310 & \textbf{\textcolor{red}{0.212}} & \textbf{\textcolor{red}{0.298}} & 0.218 & 0.304 & 0.219 & 0.305 & 0.611 & 0.588 & 0.257 & 0.345 & 0.245 & 0.322 & 
		0.496 & 0.556 & 0.239 & 0.343 \\
		& 192 & \textbf{\textcolor{red}{0.276}} &  \textcolor{blue}{\underline{0.337}} & 0.278 & \textbf{\textcolor{red}{0.334}} & 0.289 & 0.343 &  \textcolor{blue}{\underline{0.277}} & 0.338 & 0.279 & 0.338 & 0.280 & 0.341 & 0.789 & 0.685 & 0.318 & 0.380 & 0.293 & 0.346 & 1.798 & 1.137 & 0.320 & 0.397 \\
		& 336 & \textbf{\textcolor{red}{0.334}} & \textbf{\textcolor{red}{0.367}} & 0.338 &  \textcolor{blue}{\underline{0.369}} & 0.347 & 0.376 &  \textcolor{blue}{\underline{0.336}} & 0.371 & 0.342 & 0.376 & 0.341 & 0.376 & 1.469 & 0.927 & 0.375 & 0.417 & 0.361 & 0.382 & 2.929 & 1.472 & 0.409 & 0.453 \\
		& 720 & \textbf{\textcolor{red}{0.426}} & \textbf{\textcolor{red}{0.416}} &  \textcolor{blue}{\underline{0.431}} &  \textcolor{blue}{\underline{0.418}} & 0.450 & 0.430 & 0.435 & 0.424 & 0.431 & 0.419 & 0.432 & 0.426 & 1.612 & 0.957 & 0.480 & 0.472 & 0.460 & 0.432 & 4.489 & 1.782 & 0.629 & 0.565 \\
		& avg & \textbf{\textcolor{red}{0.313}} & \textbf{\textcolor{red}{0.355}} & 0.316 &  \textcolor{blue}{\underline{0.355}} & 0.329 & 0.365 &  \textcolor{blue}{\underline{0.315}} & 0.357 & 0.317 & 0.359 & 0.318 & 0.362 & 1.120 & 0.789 & 0.357 & 0.403 & 0.339 & 0.370 & 2.428 & 1.236 & 0.399 & 0.439 \\
		
		\midrule
		
		\multirow{5}{*}{\textbf{h2 → m1}}
		& 96 & 0.848 & 0.575 & 0.897 & 0.589 & 1.295 & 0.729 & 0.891 & 0.587 & 0.985 & 0.604 & 0.815 & \textbf{\textcolor{red}{0.560}} & 1.032 & 0.620 & \textbf{\textcolor{red}{0.734}} & 0.578 & 1.205 & 0.678 &  \textcolor{blue}{\underline{0.743}} & 0.577 & 0.762 &  \textcolor{blue}{\underline{0.567}} \\
		& 192 & 0.863 & 0.602 & 0.864 &  \textcolor{blue}{\underline{0.584}} & 1.196 & 0.701 & 0.850 & \textbf{\textcolor{red}{0.583}} & 0.872 & 0.600 & 0.900 & 0.606 & 1.176 & 0.676 & \textbf{\textcolor{red}{0.723}} & 0.594 & 1.159 & 0.670 &  \textcolor{blue}{\underline{0.750}} & 0.588 & 0.785 & 0.588 \\
		& 336 & 0.849 & 0.604 & 0.816 & \textbf{\textcolor{red}{0.585}} & 1.087 & 0.670 & 0.853 & 0.594 & 0.926 & 0.614 & 0.906 & 0.602 & 1.199 & 0.718 & \textbf{\textcolor{red}{0.750}} &  \textcolor{blue}{\underline{0.590}} & 1.197 & 0.689 &  \textcolor{blue}{\underline{0.764}} & 0.606 & 0.767 & 0.594 \\
		& 720 & 0.932 & 0.634 &  \textcolor{blue}{\underline{0.768}} & \textbf{\textcolor{red}{0.589}} & 0.770 &  \textcolor{blue}{\underline{0.590}} & 0.879 & 0.616 & 0.899 & 0.624 & 0.866 & 0.619 & 1.373 & 0.832 & \textbf{\textcolor{red}{0.760}} & 0.592 & 1.583 & 0.784 & 0.801 & 0.634 & 0.800 & 0.627 \\
		& avg & 0.873 & 0.604 & 0.836 & \textbf{\textcolor{red}{0.586}} & 1.087 & 0.673 & 0.868 & 0.595 & 0.920 & 0.610 & 0.871 & 0.596 & 1.195 & 0.711 & \textbf{\textcolor{red}{0.741}} &  \textcolor{blue}{\underline{0.588}} & 1.286 & 0.705 &  \textcolor{blue}{\underline{0.764}} & 0.601 & 0.778 & 0.594 \\
		
		\midrule
		
		\multirow{5}{*}{\textbf{h2 → m2}}
		& 96 & \textbf{\textcolor{red}{0.220}} & \textbf{\textcolor{red}{0.300}} &  \textcolor{blue}{\underline{0.225}} &  \textcolor{blue}{\underline{0.310}} & 0.272 & 0.351 & 0.228 & 0.311 & 0.235 & 0.316 & 0.288 & 0.345 & 0.821 & 0.634 & 0.261 & 0.347 & 0.244 & 0.324 & 
		0.327 & 0.414 & 0.264 & 0.366 \\
		& 192 & \textbf{\textcolor{red}{0.281}} & \textbf{\textcolor{red}{0.339}} &  \textcolor{blue}{\underline{0.283}} & 0.342 & 0.323 & 0.376 & 0.283 &  \textcolor{blue}{\underline{0.341}} & 0.287 & 0.346 & 0.344 & 0.375 & 1.732 & 1.018 & 0.313 & 0.370 & 0.331 & 0.374 & 0.450 & 0.485 & 0.394 & 0.452 \\
		& 336 & \textbf{\textcolor{red}{0.338}} & \textbf{\textcolor{red}{0.372}} &  \textcolor{blue}{\underline{0.340}} &  \textcolor{blue}{\underline{0.373}} & 0.369 & 0.397 & 0.343 & 0.376 & 0.361 & 0.391 & 0.438 & 0.425 & 2.587 & 1.393 & 0.401 & 0.431 & 0.386 & 0.405 & 0.526 & 0.526 & 0.506 & 0.513 \\
		& 720 & \textbf{\textcolor{red}{0.428}} & 0.430 &  \textcolor{blue}{\underline{0.429}} & \textbf{\textcolor{red}{0.418}} & 0.432 &  \textcolor{blue}{\underline{0.418}} & 0.437 & 0.424 & 0.444 & 0.433 & 0.611 & 0.588 & 3.034 & 1.452 & 0.487 & 0.472 & 0.485 & 0.458 & 0.806 & 0.652 & 0.822 & 0.655 \\
		& avg & \textbf{\textcolor{red}{0.317}} & \textbf{\textcolor{red}{0.360}} &  \textcolor{blue}{\underline{0.319}} &  \textcolor{blue}{\underline{0.360}} & 0.349 & 0.386 & 0.322 & 0.363 & 0.331 & 0.371 & 0.420 & 0.433 & 2.043 & 1.124 & 0.365 & 0.405 & 0.361 & 0.390 & 0.527 & 0.519 & 0.496 & 0.496 \\
		
		
		\bottomrule[1.2pt]
	\end{tabular}
\end{table*}

\begin{table*}[h]
	\centering
	\caption{Dataset descriptions. \textit{Variables} denotes the dimension of the multivariate time series. \textit{Frequency} indicates the sampling interval.}
	\label{tab:dataset_stat}
	\resizebox{\linewidth}{!}{
		\begin{tabular}{c|c|c|c|c|c}
			\toprule
			\textbf{Dataset} & \textbf{Variables} & \textbf{Frequency} & \textbf{Scope} & \textbf{Time Range}  & \textbf{Predicted Window}\\
			\midrule
			ETTh1, ETTh2 & 7 & 1 Hour & Energy & 2016/07 -- 2018/07 &\{96,192,336,720\} \\
			ETTm1, ETTm2 & 7 & 15 Mins & Energy & 2016/07 -- 2018/07 &\{96,192,336,720\}\\
			Electricity & 321 & 1 Hour & Energy & 2012 -- 2014 &\{96,192,336,720\}\\
			Weather & 21 & 10 Mins & Weather & 2020 Whole Year &\{96,192,336,720\}\\
			Exchange & 8 & 1 Day & Finance & 1990 -- 2016&\{96,192,336,720\} \\
			\bottomrule
		\end{tabular}
	}
\end{table*}
\begin{figure*}[h]   
	\begin{minipage}[b]{1\textwidth} 
		\includegraphics[width=\textwidth]{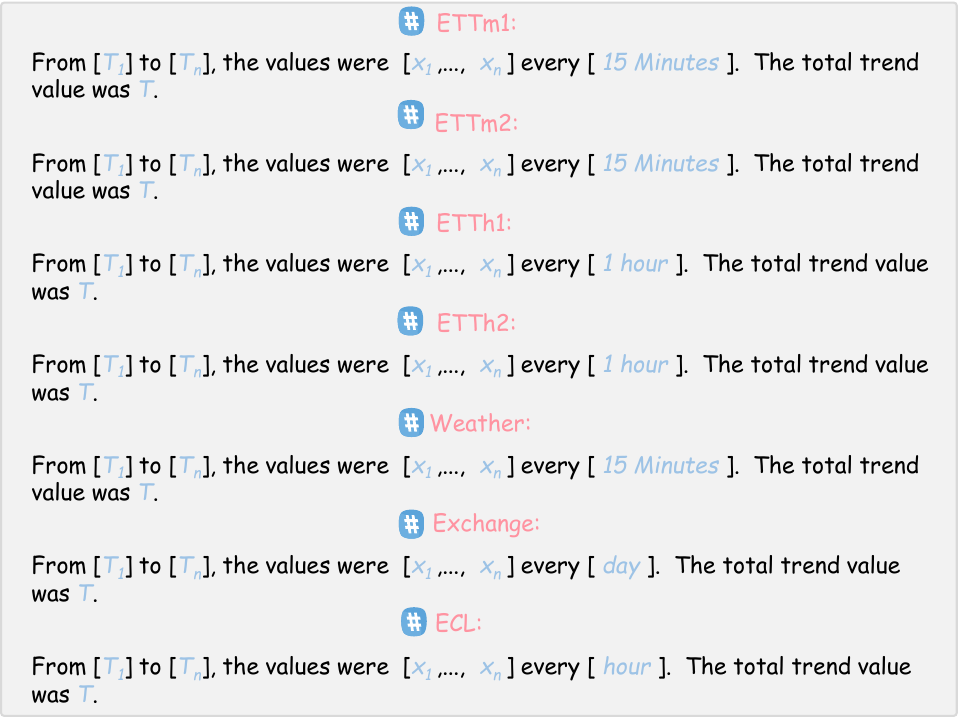}
		\centering
		
	\end{minipage}
	\caption{Hint templates for specific datasets are used to transcribe multivariate time series segments into natural language descriptions. Each template jointly encodes the time interval, numerical sequence, sampling frequency, and trend statistics to establish a unified representation between structured time series data and language model input.}
	\label{fig:llmpro}
\end{figure*}

\section{Implementation Details}\label{app:details}
Our model is implemented in Python 3.10 with the PyTorch 2.2 framework. All training and inference are conducted on a compute cluster equipped with eight NVIDIA GeForce RTX 3090 GPUs. We adopt the Adam optimizer and perform grid search over learning rates in $\{1 \times 10^{-4}, 3 \times 10^{-4}, 5 \times 10^{-4}, 1 \times 10^{-3}\}$ , while the batch size is selected from $\{16, 32, 48, 64\}$.   For key architectural hyperparameters, we conduct extensive tuning to determine the final configuration: (1) the total number of layers in the unimodal backbones is chosen from $\{2, 4\}$ ; (2) the number of fusion layers is selected from $\{1, 2, 4\}$ ; and (3) the model dimension \textit{D} is searched over $\{64, 128, 256, 512\}$ . To ensure reproducibility, all experiments use a fixed random seed of 2024. For baseline models, to maintain fairness and consistency, the first two baselines are re-implemented using their official code under the same experimental environment, UniTime \cite{liu2024unitime} results are directly taken from its original paper, and the remaining baselines follow the reported results in the iTransformer \cite{15} and CALF \cite{liu2025calf} papers.
\subsection{Evaluation Metrics}
We choose mean square error and mean absolute error as the commonly used performance evaluation indicators in time series forecasting. Their mathematical definitions are as follows:
\begin{equation}\label{eq:16}
	\begin{aligned}
		\begin{array}{l}
			MSE = \frac{1}{H}\sum\limits_{i = 1}^H {{{({{ \textbf{Y}}_i} - {{\hat { \textbf{Y}}}_i})}^2}} \\
			MAE = \frac{1}{H}\sum\limits_{i = 1}^H {\left| {{{ \textbf{Y}}_i} - {{\hat { \textbf{Y}}}_i}} \right|} 
		\end{array}
	\end{aligned}
\end{equation}

where \({{ \textbf{Y}}_i}\) denotes the true value, \({{\hat { \textbf{Y}}}_i}\) is the predicted value, and \textit{H} denotes the size of the prediction window.

\section{Prompt Description}\label{app:llm}
For all experiments, we convert each multivariate time-series window into a short natural-language description before feeding it into the frozen GPT-2 encoder. As shown in Fig. \ref{fig:llmpro}. For a given variable, we instantiate the following template:
\begin{quote}
	From [T$_1$] to [T$_n$], the values were [x$_1$, \dots, x$_n$] every [$ f$]. The total trend value was [T].
\end{quote}
Here [T$_1$] and [T$_n$] denote the start and end timestamps of the window, [x$_i$] are the observed values sampled at the dataset-specific resolution $\Delta t$ (15 minutes for ETTm1/ETTm2, 1 hour for ETTh1/ETTh2/ECL, 10 minutes for Weather, and 1 day for Exchange), and [T] is a scalar trend statistic over the window. This template is applied independently to each variable, and the resulting tokenized prompts are padded and stacked along the variable dimension to construct the textual input tensor used in our model.

\section{Theoretical Rationale of Hierarchical Asynchronous Fusion}
\label{app:theory}

In this section we give a stylized variance analysis to explain why deep layer-wise semantic coupling is more sensitive to semantic noise than the proposed hierarchical asynchronous fusion. The goal is not to provide a strict generalization bound, but to show how semantic noise can accumulate with depth in a simplified setting.

\subsection{Simplified Setup}

For clarity, we consider one scalar feature dimension and linearize the forward propagation around a fixed point. Let $h_l$ denote the hidden state at layer $l$, and let $F(\cdot)$ summarize the deterministic transformation of self-attention and FFN. We assume that at each fusion operation, the semantic branch provides a signal that can be decomposed as
\[
s_l = \mu + \varepsilon_l ,
\]
where $\mu$ is the useful semantic component shared across layers and $\varepsilon_l$ is zero-mean semantic noise (including mismatch between text and time series as well as structural noise), with
\[
\mathbb{E}[\varepsilon_l] = 0, \qquad \mathrm{Var}(\varepsilon_l) \le \sigma^2 .
\]

We emphasize that this analysis is purely illustrative and relies on linearization and independence assumptions; it is intended to clarify the intuition behind hierarchical asynchronous fusion rather than to serve as a rigorous guarantee for the full non-linear model.
\subsection{Noise Accumulation in Deep Synchronous Fusion}

In deep synchronous fusion, semantic information is injected at {every} layer. A linearized update can be written as
\begin{equation}
	h^{\text{syn}}_{l+1}
	\approx F\big(h^{\text{syn}}_l\big) + \lambda (\mu + \varepsilon_l) ,
	\label{eq:syn_update}
\end{equation}
where $\lambda$ controls the injection strength. Unrolling $L$ layers gives
\begin{equation}
	h^{\text{syn}}_L
	\approx h_0 + \sum_{l=0}^{L-1} F\big(h^{\text{syn}}_l\big)
	+ \sum_{l=0}^{L-1} \lambda (\mu + \varepsilon_l).
	\label{eq:syn_unroll}
\end{equation}
The noise accumulated from the semantic branch is
\begin{equation}
	E_{\text{syn}}
	= \sum_{l=0}^{L-1} \lambda \varepsilon_l .
	\label{eq:Esyn_def}
\end{equation}
Its variance is
\begin{equation}
	\mathrm{Var}(E_{\text{syn}})
	= \lambda^2 \, \mathrm{Var}\!\left(\sum_{l=0}^{L-1} \varepsilon_l\right).
\end{equation}
By Cauchy–Schwarz,
\begin{equation}
	\mathrm{Var}\!\left(\sum_{l=0}^{L-1} \varepsilon_l\right)
	\le \Big(\sum_{l=0}^{L-1} \sqrt{\mathrm{Var}(\varepsilon_l)}\Big)^2
	\le L^2 \sigma^2 ,
\end{equation}
thus we obtain the following upper bound:
\begin{equation}
	\mathrm{Var}(E_{\text{syn}})
	\le L^2 \lambda^2 \sigma^2 .
	\label{eq:Esyn_var}
\end{equation}
This shows that, in the worst case, the variance of semantic noise in deep synchronous fusion can grow {quadratically} with the network depth $L$. When the per-layer noises are positively correlated (which is plausible since they are generated from the same prompt and fusion mechanism), this bound can be nearly tight.

\subsection{Noise Accumulation in Hierarchical Asynchronous Fusion}

In TimeSAF, semantic information is injected only at a small set of fusion layers. Suppose the backbone has depth $L$ and there are $S$ fusion stages, with fusion indices collected in a set $\mathcal{K}_{\text{fusion}}$. For each fusion stage $s \in \mathcal{K}_{\text{fusion}}$, a linearized update of the temporal branch can be written as
\begin{equation}
	h^{\text{asy}}_{\kappa_s+1}
	\approx F\big(h^{\text{asy}}_{\kappa_s}\big)
	+ \lambda_s (\mu + \varepsilon_s),
	\label{eq:asy_update}
\end{equation}
where $\lambda_s$ denotes the semantic injection strength at stage $s$. Semantic noise is now accumulated only at these $S$ layers:
\begin{equation}
	E_{\text{asy}}
	= \sum_{s \in \mathcal{K}_{\text{fusion}}} \lambda_s \varepsilon_s .
	\label{eq:Easy_def}
\end{equation}
Similarly,
\begin{equation}
\begin{array}{l}
	{\rm{Var}}({E_{{\rm{asy}}}}) = {\rm{Var}}\left( {\sum\limits_{s \in {{\cal K}_{{\rm{fusion}}}}} {{\lambda _s}} {\varepsilon _s}} \right)\\
	\le {(\sum\limits_{s \in {{\cal K}_{{\rm{fusion}}}}} | {\lambda _s}|\sqrt {{\rm{Var}}({\varepsilon _s})} )^2}\\
	\le {S^2}{\lambda _{{{\max }^2}}}{\sigma ^2}
\end{array}
	\label{eq:Easy_var}
\end{equation}
where $\lambda_{\max} = \max_s |\lambda_s|$. If we assume that the injection strengths of the two schemes are comparable, i.e., $\lambda_{\max} \approx \lambda$, then combining \eqref{eq:Esyn_var} and \eqref{eq:Easy_var} yields
\begin{equation}
	\frac{\mathrm{Var}(E_{\text{asy}})}{\mathrm{Var}(E_{\text{syn}})}
	\;\lesssim\;
	\frac{S^2}{L^2}
	\ll 1 ,
\end{equation}
because in practice $S \ll L$ (e.g., $S=2$ vs.\ $L=6$ in our experiments). Moreover, TimeSAF introduces a learnable gating factor $\sigma(g) \in [0,1]$ on each refinement connection, which effectively scales down $\lambda_s$ and further suppresses semantic noise injection when prompts are uninformative. This provides an additional safeguard against semantic perceptual dissonance.

\subsection{Discussion}

The above analysis is based on a linearized one-dimensional abstraction and ignores higher-order nonlinear effects. Nevertheless, it clearly shows that repeatedly injecting noisy semantic signals at every layer can lead to much stronger noise accumulation than injecting them at a small number of carefully selected fusion stages. This provides a theoretical rationale for why the proposed hierarchical asynchronous fusion is empirically more robust than deep layer-wise semantic coupling.

\section{Algorithm Pseudocode}
\label{app:code}

Algorithm \ref{alg:timesaf} outlines the forward pass of TimeSAF using a PyTorch-like style. The core distinction of our asynchronous fusion is highlighted in the conditional execution of the fusion block.
\begin{algorithm}[t]
	\caption{Forward pass of \textbf{TimeSAF}}
	\label{alg:timesaf}
	\begin{algorithmic}[1]
		\Require Historical multivariate series $\mathbf{X} \in \mathbb{R}^{B \times L \times N}$,
		LLM-based prompts $\mathbf{E}^{LLM}$,
		fusion layer indices $\{\kappa_s\}_{s=1}^{S}$,
		refinement layer index set $\mathcal{R}$,
		parameters of TimeSAF.
		\Ensure Forecast $\hat{\mathbf{Y}} \in \mathbb{R}^{H \times N}$.
		\vspace{3pt}
		\State $\mathbf{X}_{norm} \leftarrow \text{RevIN}(\mathbf{X}, \texttt{"norm"})$
		\State $\mathcal{H}_0^{Time} \leftarrow \text{TimeSeriesEncoder}(\mathbf{X}_{norm})$ 
		\Comment{patching + projection + positional encoding}
		\State $\mathcal{H}_0^{Text} \leftarrow \text{PromptEncoder}(\mathbf{E}^{LLM})$ 
		\Comment{LLM projection + positional encoding}
		\State $s \leftarrow 1$;\quad $\mathcal{F}^{(s)} \leftarrow \texttt{None}$
		\vspace{3pt}
		\For{$\ell = 1$ to $dp$}
		\If{$\ell \in \mathcal{R}$ \textbf{and} $\mathcal{F}^{(s)} \neq \texttt{None}$}
		\Comment{asynchronous semantic refinement}
		\State $\mathcal{H}_\ell^{Time} \leftarrow 
		\text{RefiningBlock}^{Time}_\ell\big(\mathcal{H}_{\ell-1}^{Time},\, \mathcal{F}^{(s)}\big)$
		\State $\mathcal{H}_\ell^{Text} \leftarrow 
		\text{RefiningBlock}^{Text}_\ell\big(\mathcal{H}_{\ell-1}^{Text},\, \mathcal{F}^{(s)}\big)$
		\Else
		\Comment{pure unimodal encoding}
		\State $\mathcal{H}_\ell^{Time} \leftarrow 
		\text{UnimodalBlock}^{Time}_\ell(\mathcal{H}_{\ell-1}^{Time})$
		\State $\mathcal{H}_\ell^{Text} \leftarrow 
		\text{UnimodalBlock}^{Text}_\ell(\mathcal{H}_{\ell-1}^{Text})$
		\EndIf
		\vspace{2pt}
		\If{$\ell = \kappa_s$}
		\Comment{stage-wise semantic fusion (bottom-up)}
		\State $\mathcal{H}_{s,0}^F \leftarrow 
		\text{Repeat}\big(\mathcal{Q}_s^F,\, B \times N\big)$
		\State $\mathcal{F}^{(s)} \leftarrow 
		\text{FusionBlock}_s\big(\mathcal{H}_{s,0}^F,\,
		\mathcal{H}_\ell^{Time},\,
		\mathcal{H}_\ell^{Text}\big)$
		\State $s \leftarrow s + 1$
		\EndIf
		\EndFor
		\vspace{3pt}
		\State $\mathbf{Y} \leftarrow \text{OutputHead}(\mathcal{H}_dp^{Time})$ 
		\Comment{flatten patches + linear projection}
		\State $\hat{\mathbf{Y}} \leftarrow \text{RevIN}(\mathbf{Y}, \texttt{"denorm"})$
		\State \Return $\hat{\mathbf{Y}}$
	\end{algorithmic}
\end{algorithm}
\section{Prompt Variant Ablation}
\label{app:prompt_ablation}

To study whether the informativeness of prompts affects the semantic features, we conduct a prompt-variant ablation with several concise prompt formulations.
For simplicity and transferability, we avoid complex prompts that require extensive domain-specific analysis; instead, we construct textual descriptions using lightweight statistical cues.
This design enables reproducible prompt generation for new datasets without additional data-specific engineering, which is aligned with practical zero-shot evaluation.
Specifically, we compare the full prompt used in Time-SAF (\textbf{Time-SAF}) with three simplified prompt types: \textbf{Domain} (domain-only prompt),
\textbf{Timestamp} (numeric-description-only prompt), and \textbf{Instruction} (instruction-only prompt).
The results are reported in Table~\ref{tab:prompt_variant_ablation}. Overall, the concise statistic-based prompting still provides stable gains.

\begin{table}[t]
	\centering
	\caption{Prompt variant ablation (MSE; lower is better).}
	\label{tab:prompt_variant_ablation}
	\setlength{\tabcolsep}{2pt}
	\small
	\renewcommand{\arraystretch}{1.1}
	\begin{tabular}{c c cccc}
		\toprule
		Dataset & $H$ & Time-SAF & Domain & Timestamp & Instruction \\
		\midrule
		\multirow{4}{*}{ETTh2}
		& 96  & 0.286 & 0.289 & 0.292 & 0.291 \\
		& 192 & 0.370 & 0.375 & 0.384 & 0.376 \\
		& 336 & 0.391 & 0.398 & 0.390 & 0.395 \\
		& 720 & 0.425 & 0.428 & 0.426 & 0.429 \\
		\midrule
		\multirow{4}{*}{Exchange}
		& 96  & 0.078 & 0.081 & 0.085 & 0.080 \\
		& 192 & 0.170 & 0.175 & 0.178 & 0.173 \\
		& 336 & 0.312 & 0.323 & 0.319 & 0.329 \\
		& 720 & 0.801 & 0.812 & 0.802 & 0.806 \\
		\midrule
		\multirow{4}{*}{Weather}
		& 96  & 0.157 & 0.156 & 0.159 & 0.162 \\
		& 192 & 0.204 & 0.208 & 0.210 & 0.209 \\
		& 336 & 0.262 & 0.266 & 0.270 & 0.269 \\
		& 720 & 0.338 & 0.339 & 0.338 & 0.341 \\
		\bottomrule
	\end{tabular}
\end{table}
\section{Additional Cross-Dataset Zero-shot Transfer}
\label{app:zeroshot_cross_dataset}

Prior zero-shot evaluations mainly within the ETT family (e.g., h$\rightarrow$m transfers) correspond to relatively mild domain shifts, which may not sufficiently demonstrate robustness under larger domain gaps.  We add two cross-dataset zero-shot transfer settings as an initial validation (Table~\ref{tab:zeroshot_cross_dataset}): ETTm2$\rightarrow$Electricity and ETTm1$\rightarrow$Weather. As shown, under these more challenging transfer scenarios, Time-SAF achieves better overall performance than the representative semantic fusion baseline Time-FFM, suggesting that our fusion strategy remains reasonably robust under larger domain shifts. We appreciate the reviewer’s suggestion to include broader cross-domain tests, and we will further expand the evaluation with more cross-domain transfer tasks and more systematic domain-gap analyses in a subsequent version.

\begin{table}[t]
	\centering
	\small
	\caption{Cross-dataset zero-shot transfer results (lower is better).}
	\label{tab:zeroshot_cross_dataset}
	\setlength{\tabcolsep}{4.2pt}
	\renewcommand{\arraystretch}{1.1}
	\begin{tabular}{c c cc cc}
		\toprule
		Transfer Setting & $H$
		& \multicolumn{2}{c}{Time-SAF} & \multicolumn{2}{c}{Time-FFM} \\
		\cmidrule(lr){3-4}\cmidrule(lr){5-6}
		& & MSE & MAE & MSE & MAE \\
		\midrule
		\multirow{4}{*}{ETTm2$\rightarrow$Electricity}
		& 96  & 0.521 & 0.581 & 0.616 & 0.613 \\
		& 192 & 0.345 & 0.335 & 0.649 & 0.631 \\
		& 336 & 0.486 & 0.478 & 0.687 & 0.651 \\
		& 720 & 0.612 & 0.597 & 0.736 & 0.675 \\
		\midrule
		\multirow{4}{*}{ETTm1$\rightarrow$Weather}
		& 96  & 0.235 & 0.268 & 0.235 & 0.270 \\
		& 192 & 0.291 & 0.321 & 0.289 & 0.312 \\
		& 336 & 0.318 & 0.325 & 0.329 & 0.336 \\
		& 720 & 0.397 & 0.371 & 0.402 & 0.381 \\
		\bottomrule
	\end{tabular}
\end{table}

\section{Additional Ablation: Trunk-Decoder after Fusion Trunk}
\label{app:trunk_decoder}

To test whether the Cross-Modal Semantic Fusion Trunk can directly yield the final forecasts, we construct a structural variant termed \emph{Trunk-Decoder}. Specifically, we attach a lightweight decoder after the fusion trunk and use the trunk outputs to directly generate the $H$-step predictions. All other settings (data splits, input length, training objective, and hyperparameters) are kept identical to the default model (\emph{Fusion Trunk}), which follows the pathway ``fusion representation $\rightarrow$ controlled injection $\rightarrow$ prediction head.''

The results are summarized in Table~\ref{tab:trunk_decoder}. As shown, Trunk-Decoder works properly and produces reasonable forecasts, yet it is overall slightly weaker than the original Fusion Trunk pipeline across multiple datasets and horizons. This indicates that the trunk representation is indeed predictive, while the controlled injection pathway better preserves fine-grained numerical dynamics and more stably captures long-horizon structures, leading to superior overall performance.

\begin{table}[t]
	\centering
	\caption{Structural comparison between Fusion Trunk and Trunk-Decoder (MSE; lower is better).}
	\label{tab:trunk_decoder}
	\small
	\setlength{\tabcolsep}{7pt}
	\renewcommand{\arraystretch}{.7}
	\begin{tabular}{c c cc}
		\toprule
		Dataset & $H$ & Fusion Trunk & Trunk-Decoder \\
		\midrule
		\multirow{4}{*}{ETTm1}
		& 96  & 0.311 & 0.320 \\
		& 192 & 0.357 & 0.363 \\
		& 336 & 0.390 & 0.401 \\
		& 720 & 0.450 & 0.462 \\
		\midrule
		\multirow{4}{*}{Weather}
		& 96  & 0.157 & 0.162 \\
		& 192 & 0.204 & 0.213 \\
		& 336 & 0.262 & 0.266 \\
		& 720 & 0.338 & 0.340 \\
		\midrule
		\multirow{4}{*}{Electricity}
		& 96  & 0.143 & 0.150 \\
		& 192 & 0.160 & 0.162 \\
		& 336 & 0.176 & 0.177 \\
		& 720 & 0.219 & 0.228 \\
		\bottomrule
	\end{tabular}
\end{table}
\end{document}